\documentclass[sigconf,nonacm]{acmart}

\usepackage{amsmath}
\usepackage[font=small]{caption}
\usepackage{comment}
\usepackage{subcaption}
\usepackage[utf8]{inputenc}
\usepackage{grffile}
\usepackage{xcolor}

{}

\newcommand{\msr}{{\large$^\ast$}}
\newcommand{\stanford}{{\large$^\ddagger$}}
\newcommand{\nv}{{\large$^\dagger$}}
\newcommand{\ack}{{\large$^\star$}}

\title{Efficient Large-Scale Language Model Training on GPU Clusters Using Megatron-LM}
\author{
\rm{Deepak Narayanan\stanford\ack, Mohammad Shoeybi\nv, Jared Casper\nv, Patrick LeGresley\nv, \\ Mostofa Patwary\nv, Vijay Korthikanti\nv, Dmitri Vainbrand\nv, Prethvi Kashinkunti\nv, \\ Julie Bernauer\nv, Bryan Catanzaro\nv, Amar Phanishayee\msr, Matei Zaharia\stanford}
\\
\rm{\textit{\nv NVIDIA\hspace{0.02in} \stanford Stanford University\hspace{0.02in} \msr Microsoft Research}}
}
\thanks{\ack \footnotesize{Work done as an intern at NVIDIA}}

\copyrightyear{2021}
\acmYear{2021}
\setcopyright{acmlicensed}\acmConference[SC '21]{The International Conference for High Performance Computing, Networking, Storage and Analysis}{November 14--19, 2021}{St. Louis, MO, USA}
\acmBooktitle{The International Conference for High Performance Computing, Networking, Storage and Analysis (SC '21), November 14--19, 2021, St. Louis, MO, USA}
\acmPrice{15.00}
\acmDOI{10.1145/3458817.3476209}
\acmISBN{978-1-4503-8442-1/21/11}

\begin{document}

\begin{abstract}

Large language models have led to state-of-the-art accuracies across several tasks. However, training these models efficiently is challenging because: a) GPU memory capacity is limited, making it impossible to fit large models on even a multi-GPU server, and b) the number of compute operations required can result in unrealistically long training times. Consequently, new methods of model parallelism such as tensor and pipeline parallelism have been proposed. Unfortunately, naive usage of these methods leads to scaling issues at thousands of GPUs. In this paper, we show how tensor, pipeline, and data parallelism can be composed to scale to thousands of GPUs. We propose a novel interleaved pipelining schedule that can improve throughput by 10+\% with memory footprint comparable to existing approaches. Our approach allows us to perform training iterations on a model with 1 trillion parameters at 502 petaFLOP/s on 3072 GPUs (per-GPU throughput of 52\% of theoretical peak).

\end{abstract}

\maketitle

\renewcommand{\shortauthors}{}

\section{Introduction}

\begin{figure}[t!]
    \centering
    \includegraphics[keepaspectratio=1.0,width=0.95\columnwidth]{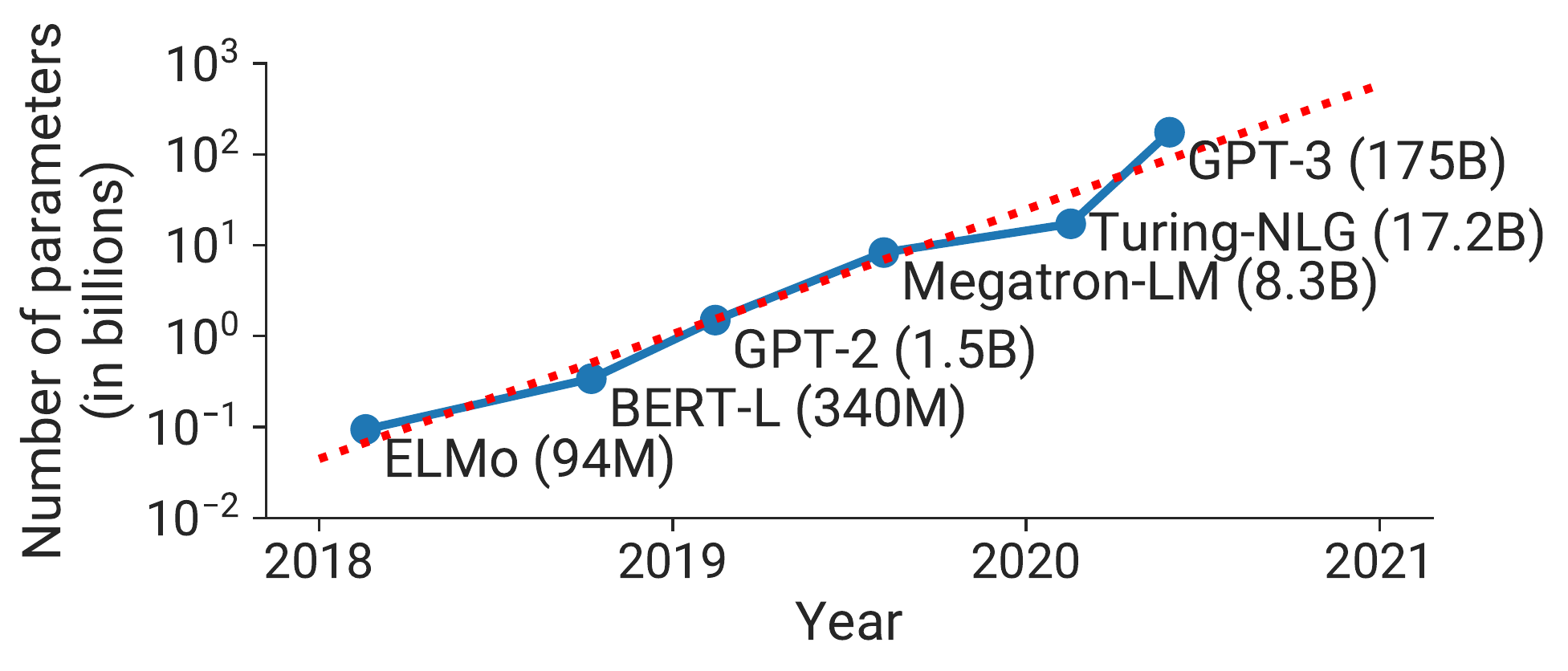}
    \vspace{-0.1in}
    \caption{
        Trend of sizes of state-of-the-art Natural Language Processing (NLP) models with time. The number of floating-point operations to train these models is increasing at an exponential rate.
    }
    \vspace{-0.1in}
    \label{fig:model_trend}
\end{figure}

Transformer-based language models~\cite{vaswani2017attention,Radford2018GPT,radford2019language,devlin2018bert, roberta,xlnet,T5} in Natural Language Processing (NLP) have driven rapid progress in recent years as computation at scale has become more available and datasets have become larger. Recent work~\cite{gpt3, shoeybi2019megatron} has shown large language models to be effective zero- or few-shot learners, with high accuracy on many NLP tasks and datasets. These large language models have a number of exciting downstream applications such as client feedback summarization, automatic dialogue generation, semantic search, and code autocompletion~\cite{gpt3applications, microsoftcodecompletion, githubcopilot}.
As a result, the number of parameters in state-of-the-art NLP models have grown at an exponential rate (Figure~\ref{fig:model_trend}). Training such models, however, is challenging for two reasons: (a) it is no longer possible to fit the parameters of these models in the main memory of even the largest GPU (NVIDIA recently released 80GB-A100 cards), and (b) even if we are able to fit the model in a single GPU (e.g., by swapping parameters between host and device memory~\cite{ren2021zero}), the high number of compute operations required can result in unrealistically long training times (e.g., training GPT-3 with 175 billion parameters~\cite{gpt3} would require approximately 288 years with a single V100 NVIDIA GPU). This calls for parallelism. Data-parallel scale-out usually works well, but suffers from two limitations: a) beyond a point, the per-GPU batch size becomes too small, reducing GPU utilization and increasing communication cost, and b) the maximum number of devices that can be used is the batch size, limiting the number of accelerators that can be used for training.

Various model parallelism techniques have been proposed to address these two challenges. For example, recent work~\cite{mesh_tf,shoeybi2019megatron} has shown how tensor (intra-layer) model parallelism, where matrix multiplications within each transformer layer are split over multiple GPUs, can be used to overcome these limitations. Although this approach works well for models of sizes up to 20 billion parameters on NVIDIA DGX A100 servers (with 8 80GB-A100 GPUs), it breaks down for larger models. Larger models need to be split across multiple multi-GPU servers, which leads to two problems: (a) the all-reduce communication required for tensor parallelism needs to go through inter-server links, which are slower than the high-bandwidth NVLink~\cite{nvlink} available within a multi-GPU server, and (b) a high degree of model parallelism can create small matrix multiplications (GEMMs), potentially decreasing GPU utilization.

Pipeline model parallelism~\cite{narayanan2019pipedream, huang2019gpipe,narayanan2021memory, yang2021pipemare, pb2021kosson, fan2021dapple} is another technique to support the training of large models, where layers of a model are striped over multiple GPUs. A batch is split into smaller microbatches, and execution is pipelined across these microbatches. Layers can be assigned to workers in various ways, and various schedules for the forward and backward passes of inputs can be used. The layer assignment and scheduling strategy results in different performance tradeoffs. Regardless of schedule, to preserve strict optimizer semantics, optimizer steps need to be synchronized across devices, leading to a \emph{pipeline flush} at the end of every batch, where microbatches are allowed to complete execution (and no new microbatches are injected). As much as 50\% of time can be spent flushing the pipeline depending on the number of microbatches injected into the pipeline. The larger the ratio of number of microbatches to the pipeline size, the smaller the time spent in the pipeline flush. Therefore, to achieve high efficiency, a larger batch size is often necessary. In this work, we also introduce a new pipeline schedule that improves efficiency at small batch sizes.

Users can thus train their large models using various techniques, each with different tradeoffs. Moreover, these techniques can be \emph{combined}. However, combining these techniques leads to non-trivial interactions, which need to be reasoned through carefully for good performance. In this paper, we address the following question:
\begin{quote}
    \emph{How should parallelism techniques be combined to maximize the training throughput of large models given a batch size while retaining strict optimizer semantics?}
\end{quote}

In particular, we show how to combine pipeline, tensor, and data parallelism, a technique we call \emph{PTD-P}, to train large language models with good computational performance (52\% of peak device throughput) on 1000s of GPUs. Our method leverages the combination of pipeline parallelism across multi-GPU servers, tensor parallelism within a multi-GPU server, and data parallelism, to practically train models with a trillion parameters with graceful scaling in an optimized cluster environment with high-bandwidth links between GPUs on the same server and across servers. We can use similar ideas to train larger models as well, given more training resources. In our experiments, we demonstrate close to linear scaling to 3072 A100 GPUs, with an achieved end-to-end training throughput of 163 teraFLOP/s per GPU (including communication, data processing, and optimization), and an aggregate throughput of 502 petaFLOP/s, on a GPT model~\cite{gpt3} with a trillion parameters using mixed precision. This throughput facilitates practical training times: we estimate end-to-end training of this model to take $\sim 3$ months. We believe this is the fastest training throughput achieved for this size of model: past systems~\cite{shoeybi2019megatron, narayanan2019pipedream} cannot train such large models since they do not combine pipeline and tensor parallelism. We also compared to ZeRO~\cite{rajbhandari2019zero}, and found that our approach outperforms ZeRO-3 by $70\%$ for models with 175 and 530 billion parameters due to less cross-node communication. These models are too large to fit on a multi-GPU server.

Achieving this throughput at scale required innovation and careful engineering along multiple axes: efficient kernel implementations that allowed most of the computation to be compute-bound as opposed to memory-bound, smart partitioning of computation graphs over the devices to reduce the number of bytes sent over network links while also limiting device idle periods, domain-specific communication optimization, and fast hardware (state-of-the-art GPUs and high-bandwidth links between GPUs on the same and different servers). We are hopeful that our open-sourced software (available at \url{https://github.com/nvidia/megatron-lm}) will enable other groups to train large NLP models efficiently at scale. 

In addition, we studied the interaction between the various components affecting throughput, both empirically and analytically when possible. Based on these studies, we offer the following guiding principles on how to configure distributed training:
\begin{itemize}
    \item Different forms of parallelism interact in non-trivial ways: the parallelization strategy has an impact on the amount of communication, the compute efficiency with which kernels are executed, as well as the idle time workers spend waiting for computation due to pipeline flushes (pipeline bubbles). For example, in our experiments, we found that sub-optimal combinations of tensor and pipeline model parallelism can lead to up to 2$\times$ lower throughput, even with high-bandwidth network links between servers; tensor model parallelism is effective within a multi-GPU server, but pipeline model parallelism must be used for larger models.
    \item The schedule used for pipeline parallelism has an impact on the amount of communication, the pipeline bubble size, and memory used to store activations. We propose a novel interleaved schedule that can improve throughput by as much as 10\% compared to previously-proposed schedules~\cite{huang2019gpipe,narayanan2021memory} with comparable memory footprint.
    \item Values of hyperparameters such as microbatch size have an impact on the memory footprint, the arithmetic efficiency of kernels executed on the worker, and the pipeline bubble size. In our experiments, the optimal value of the microbatch size is problem-dependent and can increase throughput by 15\%.
    \item At scale, distributed training is communication-intensive. When training a trillion-parameter model on 3072 GPUs, our implementation used an effective bisection bandwidth of 892 GB/s for pipeline-parallel communication, and 13 TB/s for data-parallel communication. Using slower inter-node interconnects or more communication-intensive partitionings would hinder scaling performance.
\end{itemize}
We should note that we do not automatically explore the search space of parallelism strategies (such as FlexFlow~\cite{flexflow}, PipeDream~\cite{narayanan2019pipedream}, Tarnawski et al.~\cite{tarnawski2020efficient}, and DAPPLE~\cite{fan2021dapple}), but instead suggest heuristics (in \S\ref{sec:parallelization_dimensions}) that we found work well in practice.

\begin{figure*}[t!]
    \centering
    \includegraphics[keepaspectratio=1.0,width=0.9\textwidth]{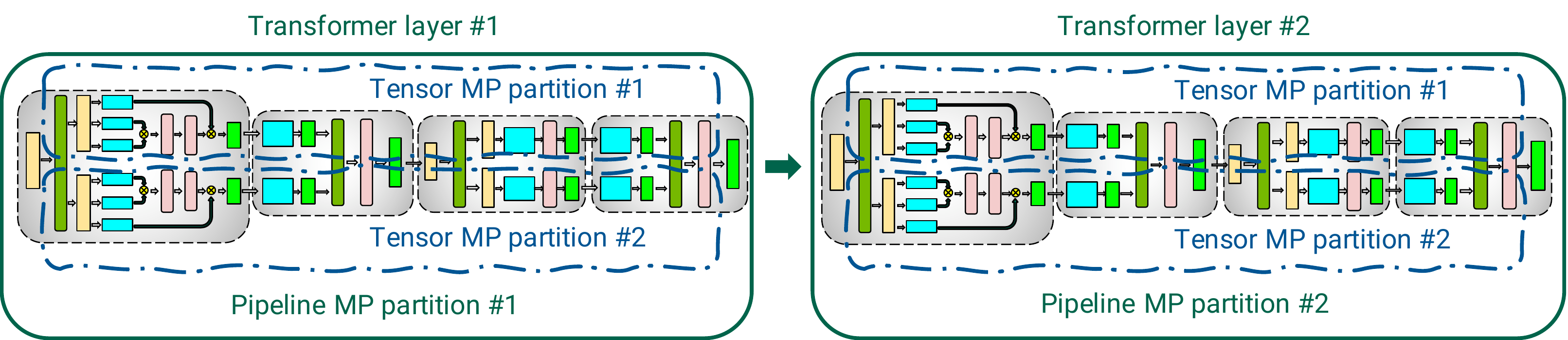}
    \vspace{-0.1in}
    \caption{
       Combination of tensor and pipeline model parallelism (MP) used in this work for transformer-based models.
    }
    \label{fig:tensor_plus_pipeline_model_parallelism}
    \vspace{-0.15in}
\end{figure*}

\begin{figure*}[t!]
    \centering
    \includegraphics[keepaspectratio=1.0,width=0.75\textwidth]{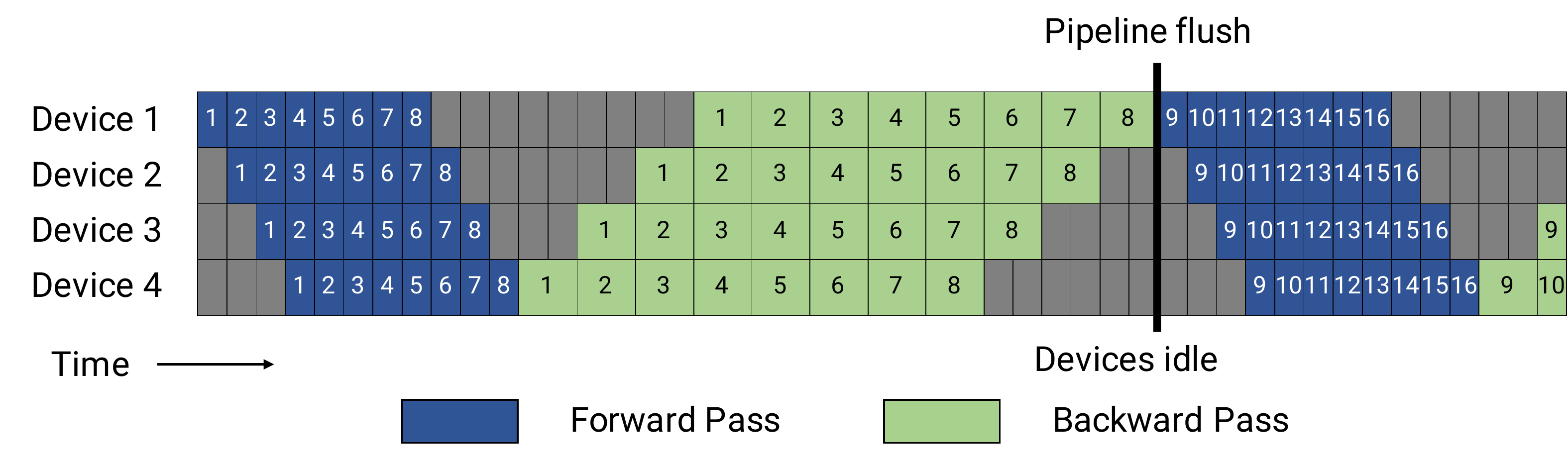}
    \vspace{-0.1in}
    \caption{
        GPipe pipeline schedule with forward passes (blue) for all microbatches (represented by numbers) followed by backward passes (green). The gray area represents the pipeline bubble. For simplicity, we assume that the backward pass takes twice as long as the forward pass. The efficiency of the pipeline schedule does not depend on this factor. Each batch in this example consists of 8 microbatches, and the numbers in each blue or green box are unique identifiers given to the corresponding microbatch (in particular, the first batch consists of microbatches $1-8$, the second batch consists of microbatches $9-16$, and so on). The optimizer is stepped and weight parameters updated at the pipeline flush to ensure strict optimizer semantics, leading to idle devices and a pipeline bubble.
    }
    \label{fig:gpipe}
\end{figure*}

\begin{figure*}[t!]
    \centering
    \includegraphics[keepaspectratio=1.0,width=0.74\textwidth]{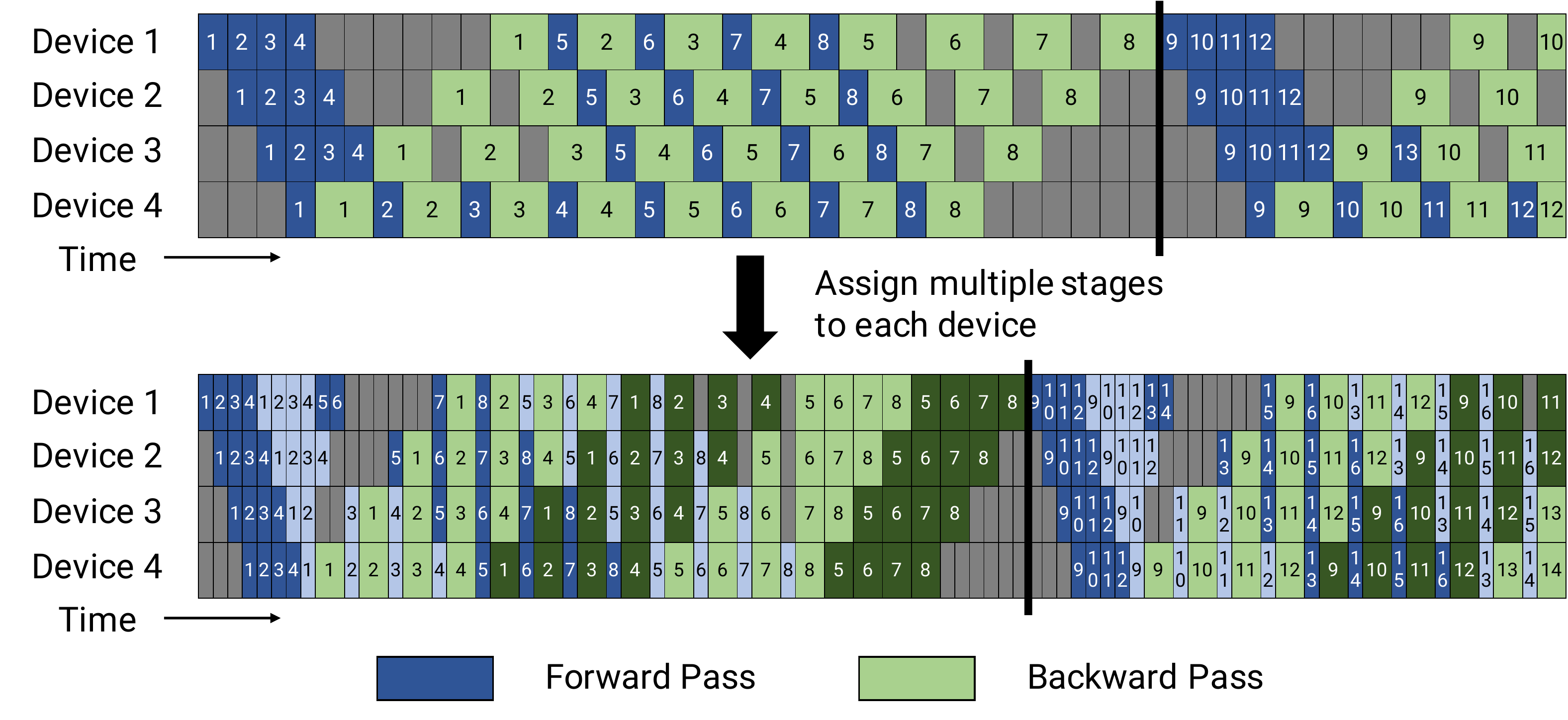}
    \vspace{-0.1in}
    \caption{
        Default and interleaved 1F1B pipeline schedules. The top figure shows the default non-interleaved 1F1B schedule. The bottom figure shows the interleaved 1F1B schedule, where each device is assigned multiple chunks (in this case, 2). Dark colors show the first chunk and light colors show the second chunk. The size of the pipeline bubble is smaller (the pipeline flush happens sooner in the interleaved timeline).
    }
    \label{fig:interleaved_schedule}
\end{figure*}
\section{Modes of Parallelism}
\label{sec:model_parallelism}

In this section, we discuss the parallelism techniques that facilitate the \emph{efficient} training of large models that do not fit in the memory of a single GPU. In this work, we combine pipeline model parallelism and tensor model parallelism (combination shown in Figure~\ref{fig:tensor_plus_pipeline_model_parallelism}) with data parallelism. We call this PTD-P for short.

\subsection{Data Parallelism}

With data parallelism~\cite{xing2015petuum, pytorchdistributed}, each worker has a copy of the full model, the input dataset is sharded, and workers aggregate their gradients periodically to ensure that all workers see a consistent version of the weights. For large models which do not fit on a single worker, data parallelism can be used on smaller model shards.

\subsection{Pipeline Model Parallelism}

With pipeline parallelism, the layers of a model are sharded across multiple devices. When used on models with the same transformer block repeated, each device can be assigned an equal number of transformer layers. We do not consider more asymmetric model architectures, where assignment of layers to pipeline stages is harder; we defer to related work~\cite{narayanan2019pipedream, flexflow, tarnawski2020efficient} to solve this problem.

A batch is split into smaller microbatches; execution is then pipelined across microbatches. Pipelining schemes need to ensure that inputs see consistent weight versions across forward and backward passes for well-defined synchronous weight update semantics. Specifically, naive pipelining can lead to an input seeing weight updates in the backward pass not seen in the forward pass.

To retain strict optimizer semantics \emph{exactly}, we introduce periodic pipeline flushes so that optimizer steps are synchronized across devices. At the start and end of every batch, devices are idle. We call this idle time the \emph{pipeline bubble}, and want to make it as small as possible. Asynchronous and bounded-staleness approaches such as PipeMare, PipeDream, and PipeDream-2BW~\cite{yang2021pipemare, pb2021kosson, narayanan2019pipedream, narayanan2021memory} do away with flushes completely, but relax weight update semantics. We defer consideration of such schemes to future work.

There are several possible ways of scheduling forward and backward microbatches across devices; each approach offers different tradeoffs between pipeline bubble size, communication, and memory footprint. We discuss two such approaches in this section.

\subsubsection{Default Schedule}
\label{sec:pipeline_parallelism_bubble}
GPipe~\cite{huang2019gpipe} proposes a schedule where the forward passes for all microbatches in a batch are first executed, followed by backward passes for all microbatches (shown in Figure~\ref{fig:gpipe}). We can quantify the size of GPipe's pipeline bubble ($t_{pb}$). We denote the number of microbatches in a batch as $m$, the number of pipeline stages (number of devices used for pipeline parallelism) as $p$, the ideal time per iteration as $t_{id}$ (assuming perfect or ideal scaling), and the time to execute a single microbatch’s forward and backward pass as $t_f$ and $t_b$. In this schedule, the pipeline bubble consists of $p-1$ forward passes at the start of a batch, and $p-1$ backward passes at the end. The total amount of time spent in the pipeline bubble is then $t_{pb}=(p-1)\cdot(t_f+t_b)$. The ideal processing time for the batch is $t_{id}=m\cdot(t_f+t_b)$. Therefore, the fraction of ideal computation time spent in the pipeline bubble is:
$$\text{Bubble time fraction (pipeline bubble size)}=\frac{t_{pb}}{t_{id}}=\frac{p-1}{m}.$$

For the bubble time fraction to be small, we thus need $m \gg p$. However, for such large $m$, this approach has a high memory footprint as it requires stashed intermediate activations (or just input activations for each pipeline stage when using activation recomputation) to be kept in memory for all $m$ microbatches through the lifetime of a training iteration.

Instead, we use the PipeDream-Flush schedule~\cite{narayanan2021memory}. In this schedule, we first enter a warm-up phase where workers perform differing numbers of forward passes as shown in Figure~\ref{fig:interleaved_schedule} (top). This schedule limits the number of in-flight microbatches (the number of microbatches for which the backward pass is outstanding and activations need to be maintained) to the depth of the pipeline, instead of the number of microbatches in a batch. After the warm-up phase, each worker then enters a steady state, where workers perform one forward pass followed by one backward pass (1F1B for short). Finally, at the end of a batch, we complete backward passes for all remaining in-flight microbatches. The time spent in the bubble is the same for this new schedule, but the number of outstanding forward passes is at most the number of pipeline stages for the PipeDream-Flush schedule. As a result, this schedule requires activations to be stashed for $p$ or fewer microbatches (compared to $m$ microbatches for the GPipe schedule). Consequently, when $m \gg p$, PipeDream-Flush is much more memory-efficient than GPipe.

\subsubsection{Schedule with Interleaved Stages}
To reduce the size of the pipeline bubble, each device can perform computation for multiple subsets of layers (called a model chunk), instead of a single contiguous set of layers. For example, if each device had 4 layers before (i.e., device 1 had layers $1-4$, device 2 had layers $5-8$, and so on), we could have each device perform computation for two model chunks (each with 2 layers), i.e., device 1 has layers $1,2,9,10$; device 2 has layers $3,4,11,12$; and so on. With this scheme, each device in the pipeline is assigned multiple pipeline stages (each pipeline stage has less computation compared to before).

As before, we can use an ``all-forward, all-backward'' version of this schedule, but this has a high memory footprint (proportional to $m$). Instead, we developed an interleaved schedule that adapts the memory-efficient 1F1B schedule from before. This new schedule is shown in Figure~\ref{fig:interleaved_schedule}, and requires the number of microbatches in a batch to be an integer multiple of the degree of pipeline parallelism (number of devices in the pipeline). For example, with 4 devices, the number of microbatches in a batch must be a multiple of 4.

As shown in Figure~\ref{fig:interleaved_schedule}, the pipeline flush for the same batch size happens sooner in the new schedule. If each device has $v$ stages (or model chunks), then the forward and backward time for a microbatch for each stage or chunk will now be $t_f/v$ and $t_b/v$. The pipeline bubble time thus reduces to $t^{\text{int.}}_{pb} = \frac{(p-1)\cdot(t_f+t_b)}{v}$, and the bubble time fraction is then:
$$\text{Bubble time fraction (pipeline bubble size)}=\frac{t^{\text{int.}}_{pb}}{t_{id}}=\frac{1}{v} \cdot \frac{p-1}{m}.$$

This means that the new schedule reduces the bubble time by $v$. This reduced pipeline bubble size, however, does not come for free: this schedule requires extra communication. Quantitatively, the amount of communication also increases by $v$. In the next section, we discuss how we can utilize the 8 InfiniBand networking cards in a multi-GPU server (e.g., a DGX A100 node) to reduce the impact of this extra communication.

\subsection{Tensor Model Parallelism}
\label{sec:tensor_model_parallelism}

\begin{figure}[t!]
 \centering
    \begin{subfigure}[c]{\columnwidth}
        \centering
        \includegraphics[keepaspectratio=1.0,width=0.85\columnwidth]{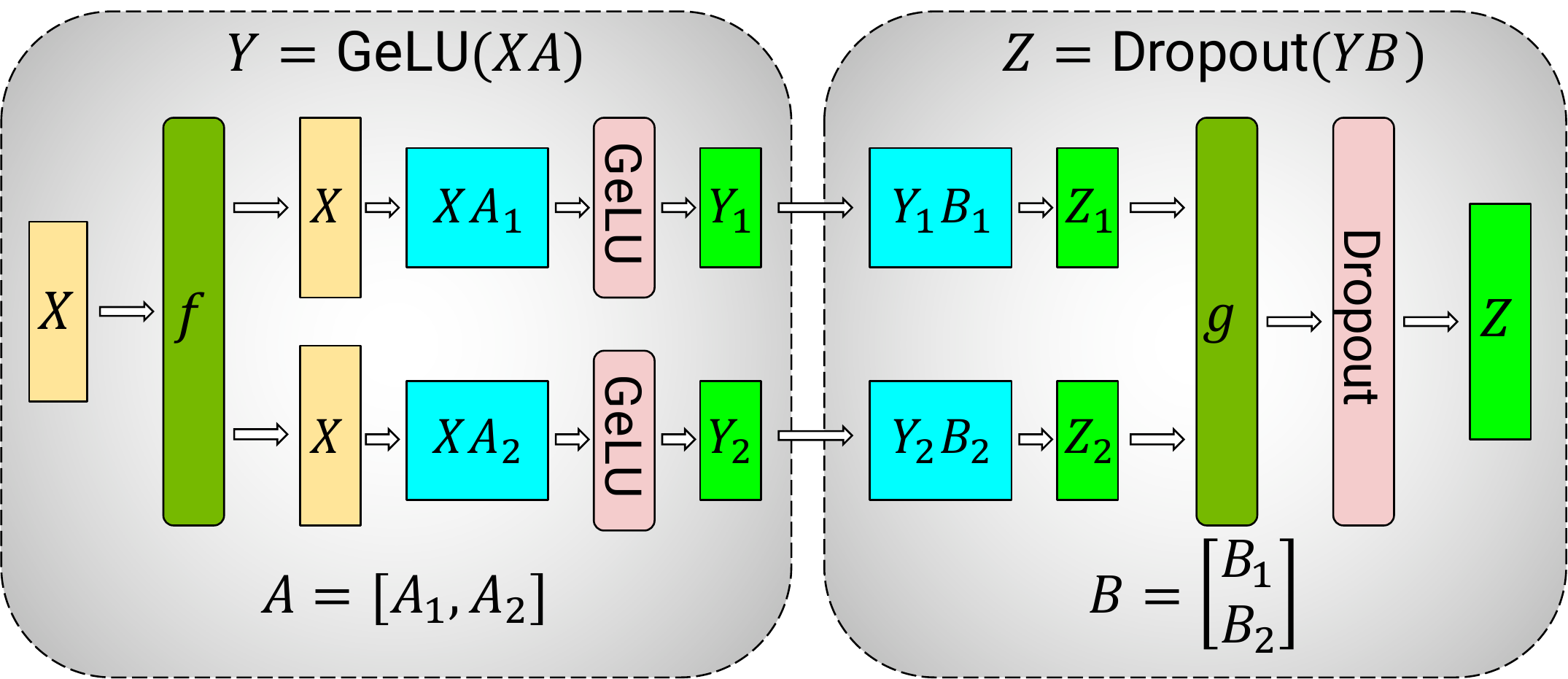}
        \caption{MLP.}
        \label{fig:tensor_model_parallelism_mlp}
    \end{subfigure}
    \begin{subfigure}[c]{\columnwidth}
        \centering
        \includegraphics[keepaspectratio=1.0,width=\columnwidth]{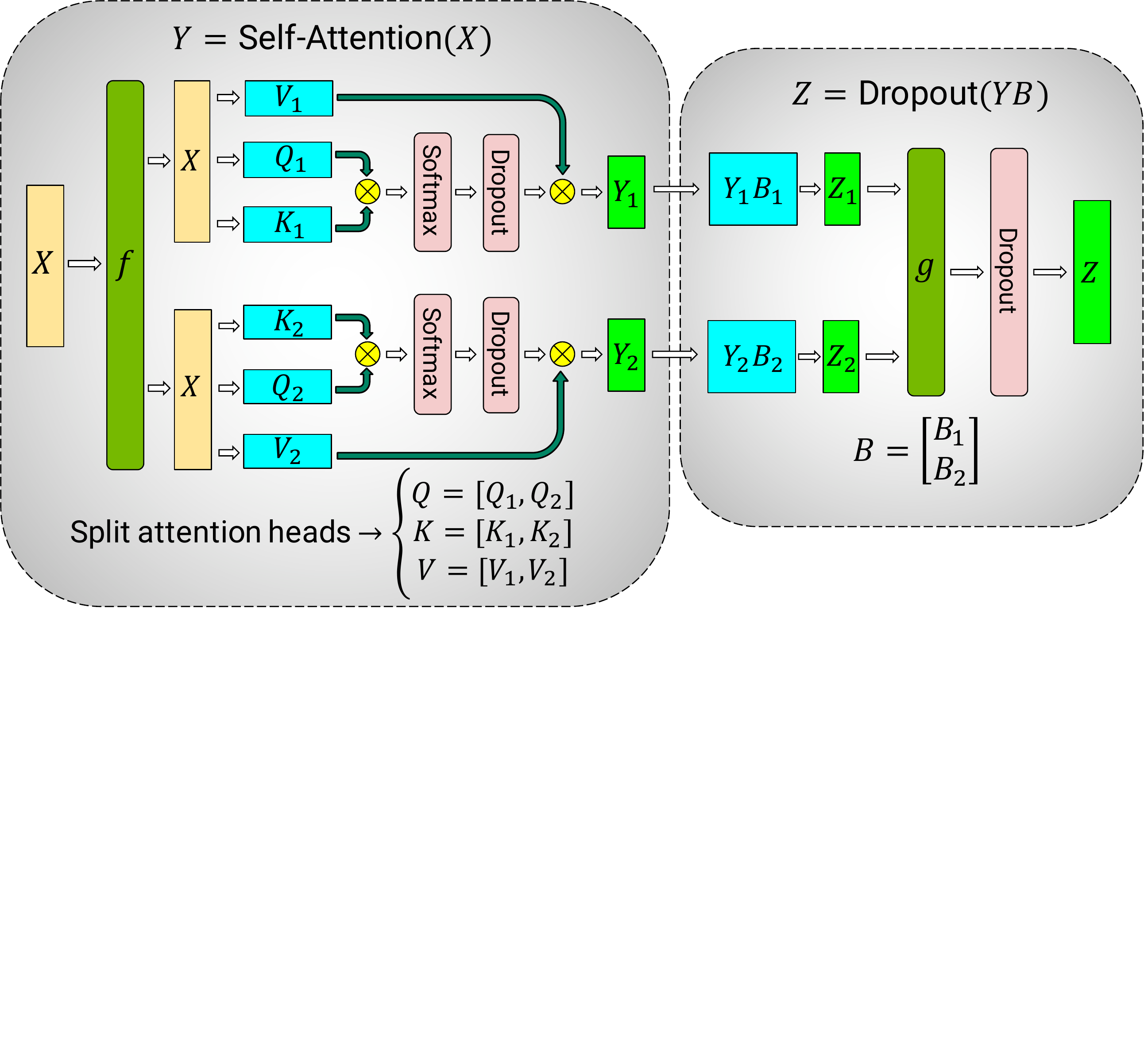}
        \vspace{-1.5in}
        \caption{Self-Attention.}
        \label{fig:tensor_model_parallelism_self_attention}
    \end{subfigure}
    \vspace{-0.1in}
    \caption{
        Blocks of transformer model partitioned with tensor model parallelism (figures borrowed from Megatron~\cite{shoeybi2019megatron}). $f$ and $g$ are conjugate. $f$ is the identity operator in the forward pass and all-reduce in the backward pass, while $g$ is the reverse.
    }
    \label{fig:tensor_model_parallelism}
    \vspace{-0.2in}
\end{figure}

With tensor model parallelism, individual layers of the model are partitioned over multiple devices. In this paper, we use the particular partitioning strategy used by Megatron~\cite{shoeybi2019megatron} for transformer layers, the bedrock of language models. We can apply similar ideas to other types of models, like CNNs, as well. We briefly outline this strategy, illustrated in Figure~\ref{fig:tensor_model_parallelism}, below.

A transformer layer consists of a self-attention block followed by a two-layer multi-layer perceptron (MLP). Further details of the transformer layer can be found in Vaswani et al~\cite{vaswani2017attention}. 

The MLP block consists of two GEMMs and a GeLU non-linearity:
$$Y = \textrm{GeLU}(XA). \, \, \, \, \, Z = \textrm{Dropout}(YB).$$
We can split $A$ along its columns $A=[A_1, A_2]$. This partitioning allows the GeLU non-linearity to be independently applied to the output of each partitioned GEMM:
 \begin{equation}
    [Y_1, Y_2]= [\textrm{GeLU}(XA_1), \textrm{GeLU}(XA_2)].  \nonumber
 \end{equation}
 This is advantageous as it removes the need for synchronization (needed if $A$ is split along its rows since GeLU is non-linear).
 
 The rows of the second weight matrix $B$ can then be split along its rows to remove the need for any communication between the GEMMs (shown in Figure~\ref{fig:tensor_model_parallelism_mlp}), as shown below:
 \begin{equation}
    B=\begin{bmatrix}
        B_1 \\
        B_2
    \end{bmatrix}, \
     Y = [Y_1, Y_2]. \nonumber
 \end{equation}
 The output of the second GEMM is then reduced across the GPUs before the dropout layer.
 
We exploit the inherent parallelism in the multi-head attention operation to partition the self-attention block (shown in Figure~\ref{fig:tensor_model_parallelism_self_attention}). The key ($K$), query ($Q$), and value ($V$) matrices can be partitioned in a column-parallel fashion. The output linear layer can then directly operate on the partitioned output of the attention operation (weight matrix partitioned across rows).

This approach splits GEMMs in the MLP and self-attention blocks across GPUs while requiring only two all-reduce operations in the forward pass ($g$ operator) and two all-reduces in the backward pass ($f$ operator). We implemented $f$ and $g$ in a few lines of code.

\section{Performance Analysis of Parallelization Configurations}
\label{sec:parallelization_dimensions}

In this section, we consider the performance implications of combining pipeline and tensor model parallelism with data parallelism. Given a fixed budget of GPUs and batch size, one can use different degrees of the parallelism types in PTD-P to train models; each dimension exposes tradeoffs between memory footprint, device utilization, and amount of communication.

We discuss these tradeoffs in the rest of this section, and then show empirical results in \S\ref{sec:evaluation_parallel_configurations}. We present analytical models where relevant for the pipeline bubble size. We qualitatively describe how communication time behaves and present cost models for amount of communication; however, we do not present direct cost models for communication time, which is harder to model for a hierarchical network topology where interconnects between GPUs on the same server have higher bandwidth than interconnects between servers. To the best of our knowledge, this is the first work to analyze the performance \emph{interactions} of these parallelization dimensions.

\subsection{Notation}
We use the following notation in this section:
\begin{itemize}
    \item $(p, t, d)$: Parallelization dimensions. $p$ for the pipeline-model-parallel size, $t$ for the tensor-model-parallel size, and $d$ for the data-parallel size.
    \item $n$: Number of GPUs. We require $p \cdot t \cdot d = n$.
    \item $B$: Global batch size (provided as input).
    \item $b$: Microbatch size.
    \item $m=\frac{1}{b}\cdot\frac{B}{d}$: Number of microbatches in a batch \emph{per pipeline}.
\end{itemize}

\subsection{Tensor and Pipeline Model Parallelism}
Tensor and pipeline model parallelism can both be used to partition a model's parameters over multiple GPUs. As stated earlier, using pipeline parallelism with periodic flushes results in a pipeline bubble of size $(p-1) / m$. Let us assume that $d=1$ (data-parallel size); consequently, $t \cdot p = n$. The pipeline bubble size in terms of $t$ is:
$$\frac{p-1}{m} = \frac{n/t-1}{m}.$$
As $t$ increases, the pipeline bubble thus decreases for fixed $B$, $b$, and $d$ ($m = B / (b \cdot d)$ is fixed as well).

The amount of communication performed between different GPUs is also affected by the values of $p$ and $t$. Pipeline model parallelism features cheaper point-to-point communication. Tensor model parallelism, on the other hand, uses all-reduce communication (two all-reduce operations each in the forward and backward pass, see \S\ref{sec:tensor_model_parallelism}). With pipeline parallelism, the total amount of communication that needs to be performed between every pair of consecutive devices (for either the forward or backward pass) for each microbatch is $bsh$, where $s$ is the sequence length and $h$ is the hidden size. With tensor model parallelism, tensors of total size $bsh$ need to be all-reduced among $t$ model replicas twice each in the forward and backward pass for each layer, leading to a total communication of $8bsh \left(\frac{t-1}{t}\right)$ \emph{per layer} per device for each microbatch. Each device typically has multiple layers; the total amount of tensor-parallel-communication per device for each microbatch is then $l^{\text{stage}} \cdot \left(8bsh \left(\frac{t-1}{t}\right)\right)$, where $l^{\text{stage}}$ is the number of layers in a pipeline stage.

Consequently, we see that tensor model parallelism increases the amount of communication between devices. Thus, when $t$ is larger than the number of GPUs in a single node, the overhead of performing tensor model parallelism across slower inter-node links can be impractical. We see these results empirically in \S\ref{sec:evaluation_parallel_configurations}.

\vspace{0.1in}
\noindent\fbox{\parbox{0.98\columnwidth}{\textbf{Takeaway \#1:} When considering different forms of model parallelism, tensor model parallelism should generally be used up to degree $g$ when using $g$-GPU servers, and then pipeline model parallelism can be used to scale up to larger models across servers.}}

\subsection{Data and Model Parallelism}
\label{sec:data_and_model_parallelism_analytical_model}

We also want to consider the interaction between data parallelism and the two types of model parallelism. In this section, we consider these interactions independently for simplicity.

\begin{figure}[t!]
    \centering
    \includegraphics[keepaspectratio=1.0,width=0.9\columnwidth]{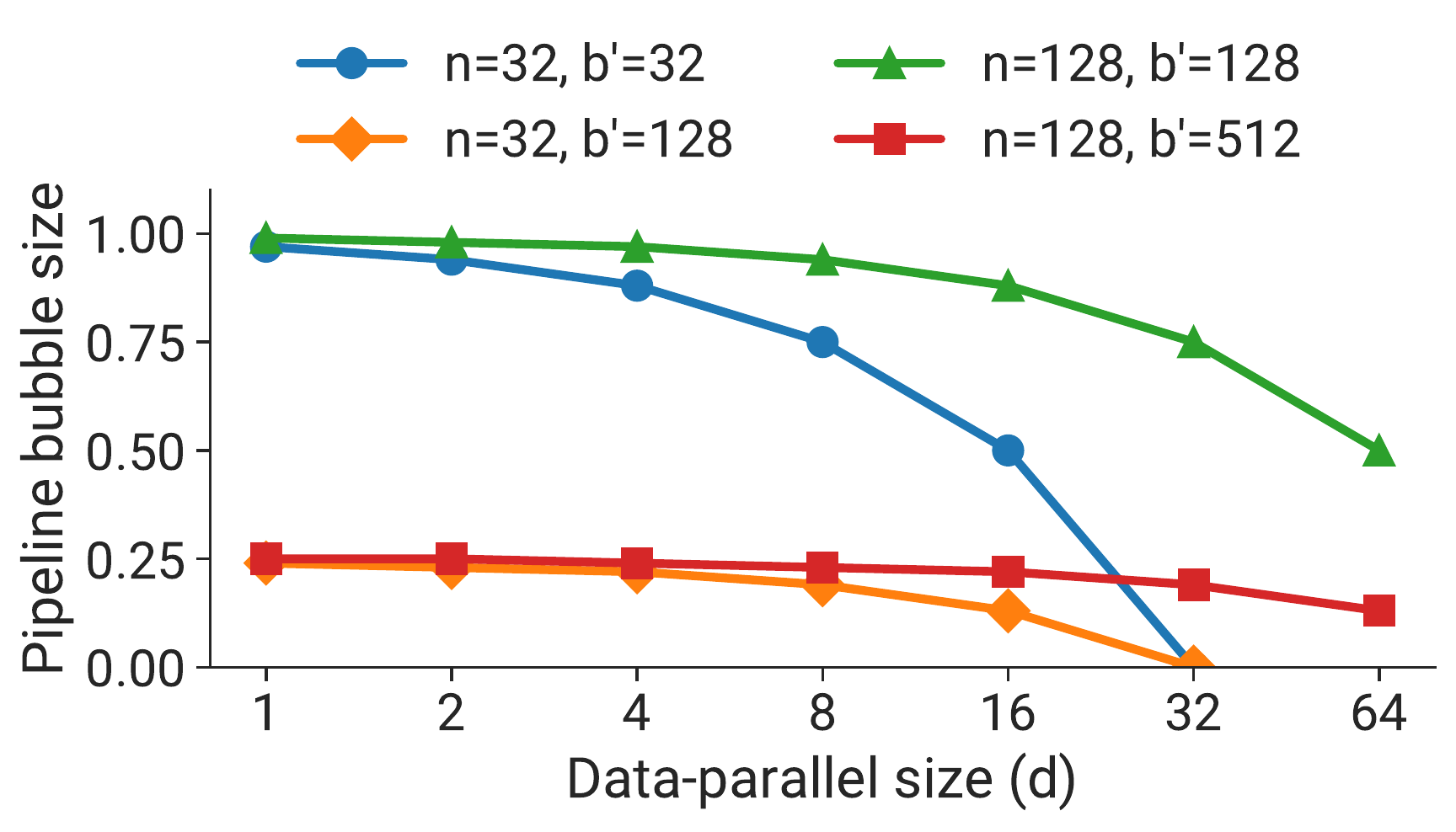}
    \vspace{-0.1in}
    \caption{
         Fraction of time spent idling due to pipeline flush (pipeline bubble size) versus data-parallel size ($d$), for different numbers of GPUs ($n$) and ratio of batch size to microbatch size ($b' = B/b$).
    }
    \vspace{-0.1in}
    \label{fig:pipeline_bubble_size_data_and_pipeline_parallelism}
\end{figure}

\subsubsection{Pipeline Model Parallelism} Let $t=1$ (tensor-model-parallel size). The number of microbatches per pipeline is $m = B / (d \cdot b)=b'/d$, where $b':=B/b$. With total number of GPUs $n$, the number of pipeline stages is $p=n/(t\cdot d)=n/d$. The pipeline bubble size is:$$\frac{p-1}{m} = \dfrac{n/d-1}{b'/d} = \dfrac{n-d}{b'}.$$
As $d$ becomes larger, $n-d$ becomes smaller, and thus the pipeline bubble becomes smaller. Figure~\ref{fig:pipeline_bubble_size_data_and_pipeline_parallelism} shows the behavior of the pipeline bubble size for various values of $d, n,$ and $b'$. It might not be possible to increase $d$ all the way to $n$ for all models, since a model's full training memory footprint might be larger than the memory capacity of a single accelerator.

Overall throughput will thus increase if the all-reduce communication needed for data parallelism does not drastically increase with higher $d$, which should hold since the communication time for a ring-based implementation scales with $\frac{d-1}{d} = 1 - \frac{1}{d}$.

We can also analyze the impact of increasing the batch size $B$. For a given parallel configuration, as the batch size $B$ increases, $b'=B/b$ increases, $(n-d)/b'$ decreases, consequently increasing throughput. All-reduce communication required by data parallelism also becomes more infrequent, further increasing throughput.

\subsubsection{Data and Tensor Model Parallelism}
With tensor model parallelism, all-reduce communication needs to be performed for \emph{every} microbatch. This can be expensive across multi-GPU servers. On the other hand, data parallelism only needs to perform expensive all-reduce communication \emph{once per batch}. Moreover, with tensor model parallelism, each model-parallel rank performs a subset of the computation in each model layer, and thus for insufficiently-large layers, modern GPUs might not perform these sub-matrix computations with peak efficiency.

\vspace{0.1in}

\noindent\fbox{\parbox{0.98\columnwidth}{\textbf{Takeaway \#2:} When using data and model parallelism, a total model-parallel size of $M = t \cdot p$ should be used so that the model's parameters and intermediate metadata fit in GPU memory; data parallelism can be used to scale up training to more GPUs.}}

\subsection{Microbatch Size}
\label{sec:microbatch_size_analytical_model}

\begin{figure}[t!]
    \centering
    \includegraphics[keepaspectratio=1.0,width=0.9\columnwidth]{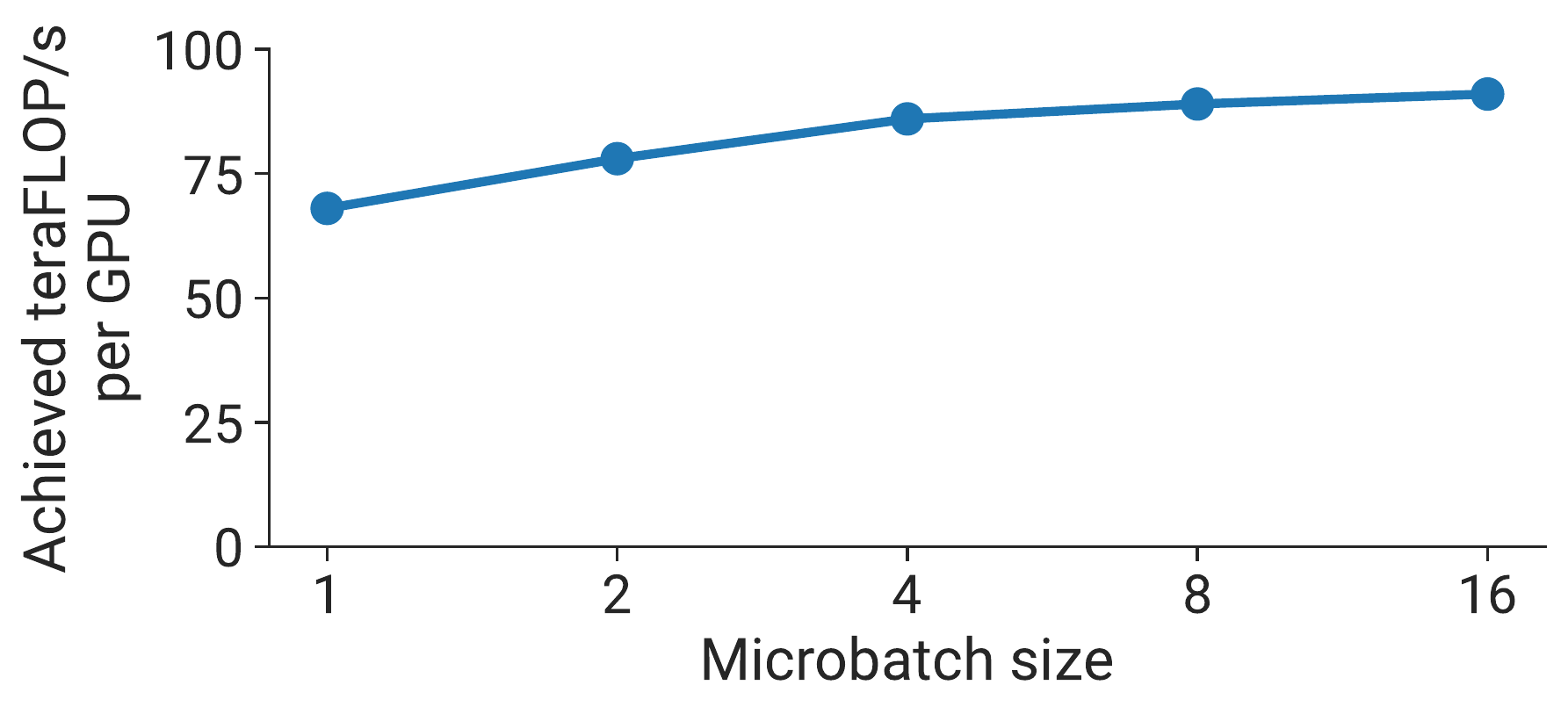}
    \vspace{-0.1in}
    \caption{
         Per-GPU throughput versus microbatch size for a GPT model with a billion parameters (128 attention heads, hidden size of 4096, 4 transformer layers).
    }
    \vspace{-0.1in}
    \label{fig:throughput_vs_microbatch_size}
\end{figure}

\begin{figure}[t!]
    \centering
    \includegraphics[keepaspectratio=1.0,width=0.9\columnwidth]{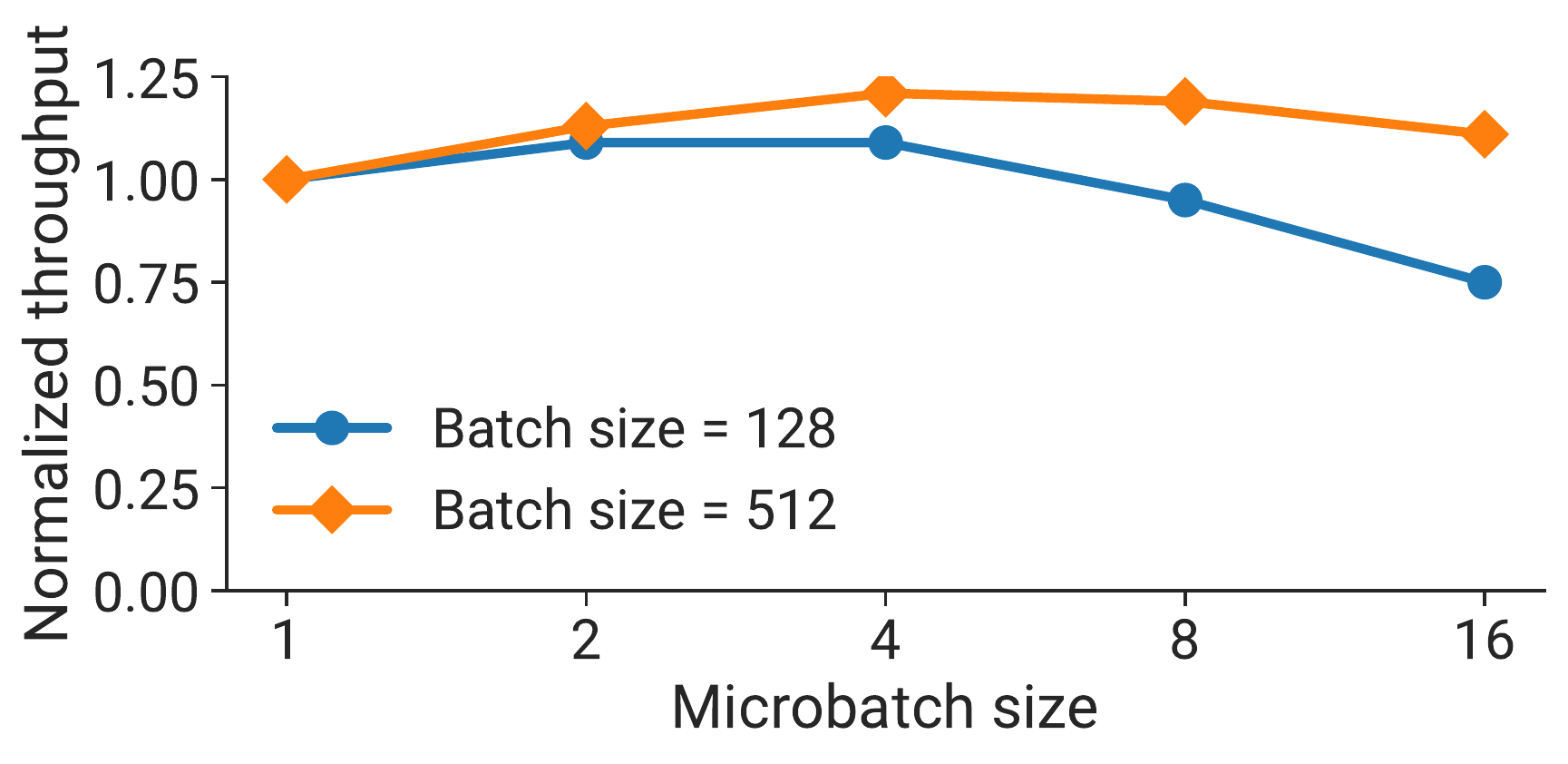}
    \vspace{-0.1in}
    \caption{
         Behavior of normalized estimated throughput  (time computed as $t = \left(b' / b + p - 1\right) \cdot \left(t_f(b) + t_b(b)\right)$) with respect to the microbatch size $b$ for the same GPT model from Figure~\ref{fig:throughput_vs_microbatch_size}.
    }
    \vspace{-0.1in}
    \label{fig:per_batch_time_estimate_vs_microbatch_size}
\end{figure}

The choice of the microbatch size $b$ also affects model-training throughput. For example, we see in Figure~\ref{fig:throughput_vs_microbatch_size} that per-GPU throughput increases by up to $1.3\times$ with a larger microbatch size on a single GPU. We now want to determine the optimal microbatch size $b$ given a parallel configuration $(p, t, d)$ and batch size $B$. The amount of data-parallel communication will be the same regardless of the microbatch size. Given functions $t_f(b)$ and $t_b(b)$ that map the microbatch size to the forward and backward computation times for a single microbatch, the total time spent computing a batch, ignoring communication cost, is (as before, define $b'$ as $B/d$):
\begin{equation}
    \left(b' / b + p - 1\right) \cdot \left(t_f(b) + t_b(b)\right). \label{equation:processing_time_estimate}
\end{equation}
The microbatch size thus affects both the arithmetic intensity of operations as well as the pipeline bubble size (by affecting $m$). Figure~\ref{fig:per_batch_time_estimate_vs_microbatch_size} shows estimated throughput (equation (\ref{equation:processing_time_estimate}) used to estimate processing time) for a GPT model with a billion parameters and $(p, t) = (8, 8)$. The optimal $b$ for both batch sizes is 4.

\vspace{0.1in}
\noindent\fbox{\parbox{0.98\columnwidth}{\textbf{Takeaway \#3:} The optimal microbatch size $b$ depends on the throughput and memory footprint characteristics of the model, as well as the pipeline depth $p$, data-parallel size $d$, and batch size $B$.}}

\subsection{Activation Recomputation}

Activation recomputation~\cite{huang2019gpipe, chen2016training, griewank2000algorithm, mlsys2020_196} is an optional technique that trades off an increase in the number of compute operations performed for additional memory footprint, by running the forward pass a second time just before the backward pass (and stashing only the input activations for a given pipeline stage, as opposed to the entire set of intermediate activations, which is much larger). Activation recomputation is required to train reasonably large models with pipeline parallelism to keep memory footprint acceptably low. Previous work like PipeDream-2BW~\cite{narayanan2021memory} has looked at the performance ramifications of activation recomputation. 

The number of activation checkpoints does not impact throughput, but impacts memory footprint. Let $A^\text{input}$ be the size of the input activations of a layer, and $A^\text{intermediate}$ be the size of intermediate activations per layer. If a model stage has $l$ layers, and if $c$ is the number of checkpoints, the total memory footprint is going to be $c \cdot A^\text{input} + l/c \cdot A^\text{intermediate}$. The minimum value of this function is obtained when $c = \sqrt{l \cdot \left(A^\text{intermediate} / A^\text{input}\right)}$. In practice, we measure $A^\text{intermediate}$ empirically. For most cases, checkpointing every 1 or 2 transformer layers is optimal.

Other techniques such as activation partitioning~\cite{rajbhandari2019zero} can also be used in conjunction with tensor model parallelsim to reduce the memory footprint due to activations further.
\section{Implementation}

We implemented PTD-P as an extension to the Megatron-LM codebase. Our implementation is built using PyTorch~\cite{pytorch}. We use NCCL~\cite{nccl} for communication between devices. To obtain good performance, we implemented optimizations targeting both communication and computation, which we outline below.
\begin{figure}[t!]
    \centering
    \begin{subfigure}[c]{0.49\columnwidth}
        \centering
        \includegraphics[keepaspectratio=1.0,width=\columnwidth]{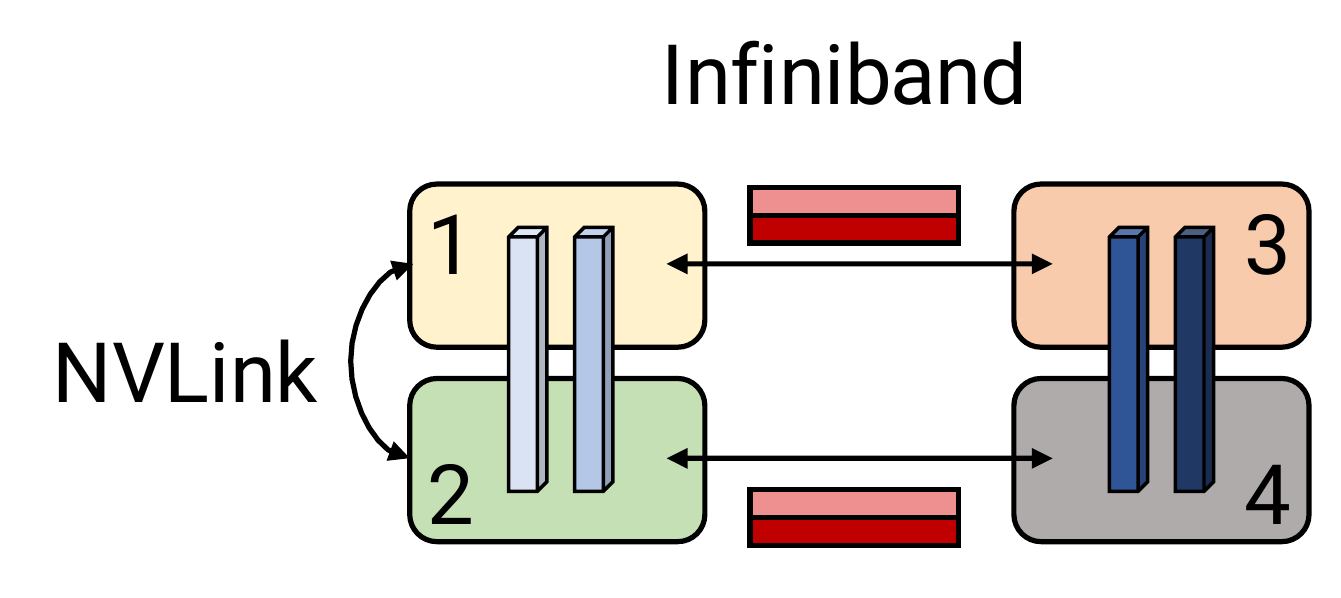}
        \caption{W/o scatter/gather optimization.}
        \label{fig:without_scatter_gather_optimization}
    \end{subfigure}
    \begin{subfigure}[c]{0.49\columnwidth}
        \centering
        \includegraphics[keepaspectratio=1.0,width=0.78\columnwidth]{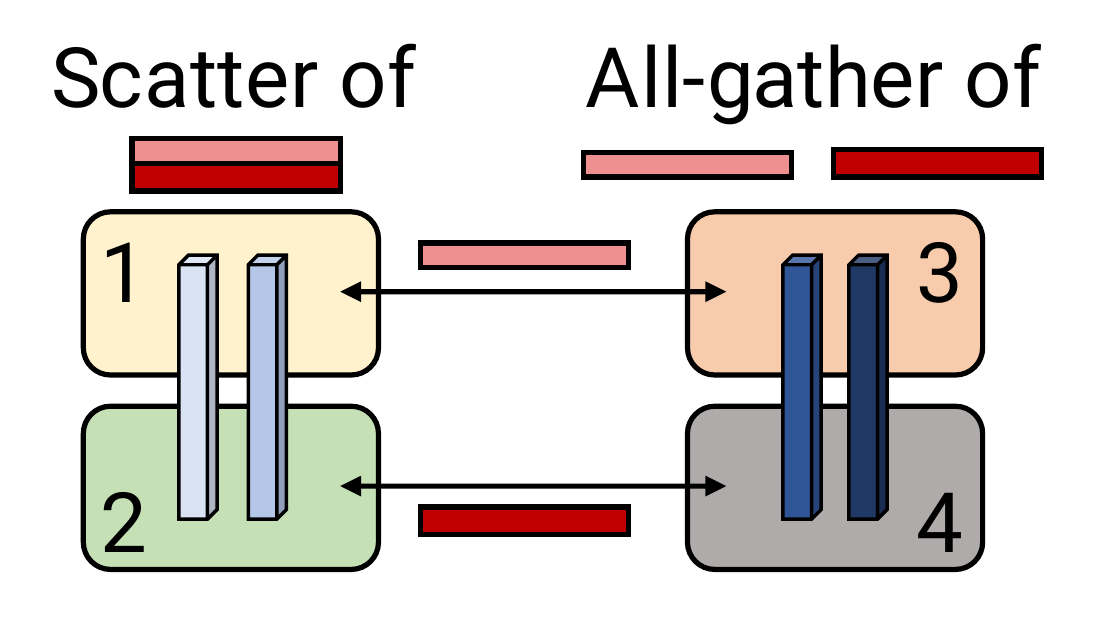}
        \caption{With scatter/gather optimization.}
        \label{fig:with_scatter_gather_optimization}
    \end{subfigure}
    \vspace{-0.1in}
    \caption{
        Scatter/gather communication optimization. Light blue blocks are layers in the first pipeline stage, and dark blue blocks are layers in the second pipeline stage. Without the scatter/gather optimization, the same tensor is sent redundantly over inter-node InfiniBand links. Instead, at the sender, we can scatter the tensor into smaller chunks, reducing the sizes of tensors sent over InfiniBand links. The final tensor can then be rematerialized at the receiver using a gather operation.
    }
    \label{fig:scatter_gather_optimization}
\end{figure}

\subsection{Communication Optimizations}

When using pipeline parallelism, we want to send and receive tensors in the forward and backward direction in parallel. Each DGX A100 is equipped with 8 InfiniBand (IB) networking cards. Unfortunately, sends and receives are point-to-point, and only happen between a pair of GPUs on two servers, making it hard to leverage all 8 cards for a single communication call within the pipeline.

However, we can leverage the fact that we use both tensor model parallelism and pipeline model parallelism to reduce the overhead of cross-node communication. In particular, we note that the output of each transformer layer is replicated (after $g$ in MLP block, see Figure \ref{fig:tensor_model_parallelism_mlp}) across the tensor-parallel ranks. As a result, ranks in two consecutive pipeline stages that are performing tensor model parallelism send and receive the exact same set of tensors (Figure~\ref{fig:without_scatter_gather_optimization}).

For large enough models, we use a tensor-model-parallel size of 8. This means we are sending the same set of tensors 8 times between corresponding GPUs on adjacent multi-GPU servers. To reduce this redundancy, we can instead split the tensor on the send side into equal-sized chunks, and then only send one chunk to the corresponding rank on the next node using the rank's own InfiniBand card (e.g., rank 1 sends to rank 3 and rank 2 sends to rank 4 in Figure~\ref{fig:scatter_gather_optimization}). With 8 tensor-model-parallel ranks, each chunk would be one-eighth smaller. Then, on the receive side, we can perform an all-gather over NVLink, which is much faster than the InfiniBand interconnect, to re-materialize the full tensor. This is shown in Figure~\ref{fig:with_scatter_gather_optimization}. We call this the \emph{scatter/gather communication optimization}. This optimization helps better leverage the multiple IB cards on the DGX A100 servers, and makes more communication-intensive schedules such as the interleaved one feasible.

Quantitatively, with the scatter-gather communication optimization, the total amount of communication that needs to be performed between every pair of consecutive stages is reduced to $\frac{bsh}{t}$, where $t$ is the tensor-model-parallel size, $s$ is the sequence length, and $h$ is the hidden size ($t=8$ in our experiments).

\subsection{Computation Optimizations}
\label{sec:computation_optimizations}

We implemented three model-specific optimizations to the computation graph to attain high performance. First, we changed the data layout in the transformer layer to avoid memory-intensive transpose operations, and to enable the use of strided batched GEMM kernels. Specifically, we changed the data layout from $[b, s, a, h]$ to $[s, b, a, h]$, where $b$, $s$, $a$, and $h$ are batch, sequence, attention-head, and hidden-size dimensions, respectively. Second, we generated fused kernels for a sequence of element-wise operations (bias + GeLU and bias + dropout + add) using PyTorch JIT~\cite{pytorchjit}. Third, we created two custom kernels to enable the fusion of scale, mask, and softmax (reduction) operations: one to support general masking (used in models such as BERT) and another to support implicit causal masking (used in auto-regressive models such as GPT). We quantify the effect of these optimizations in the next section.

\section{Evaluation}

\begin{table*}[t!]
    \vspace{-0.5in}
    \centering
    \includegraphics[keepaspectratio=1.0,width=0.97\textwidth]{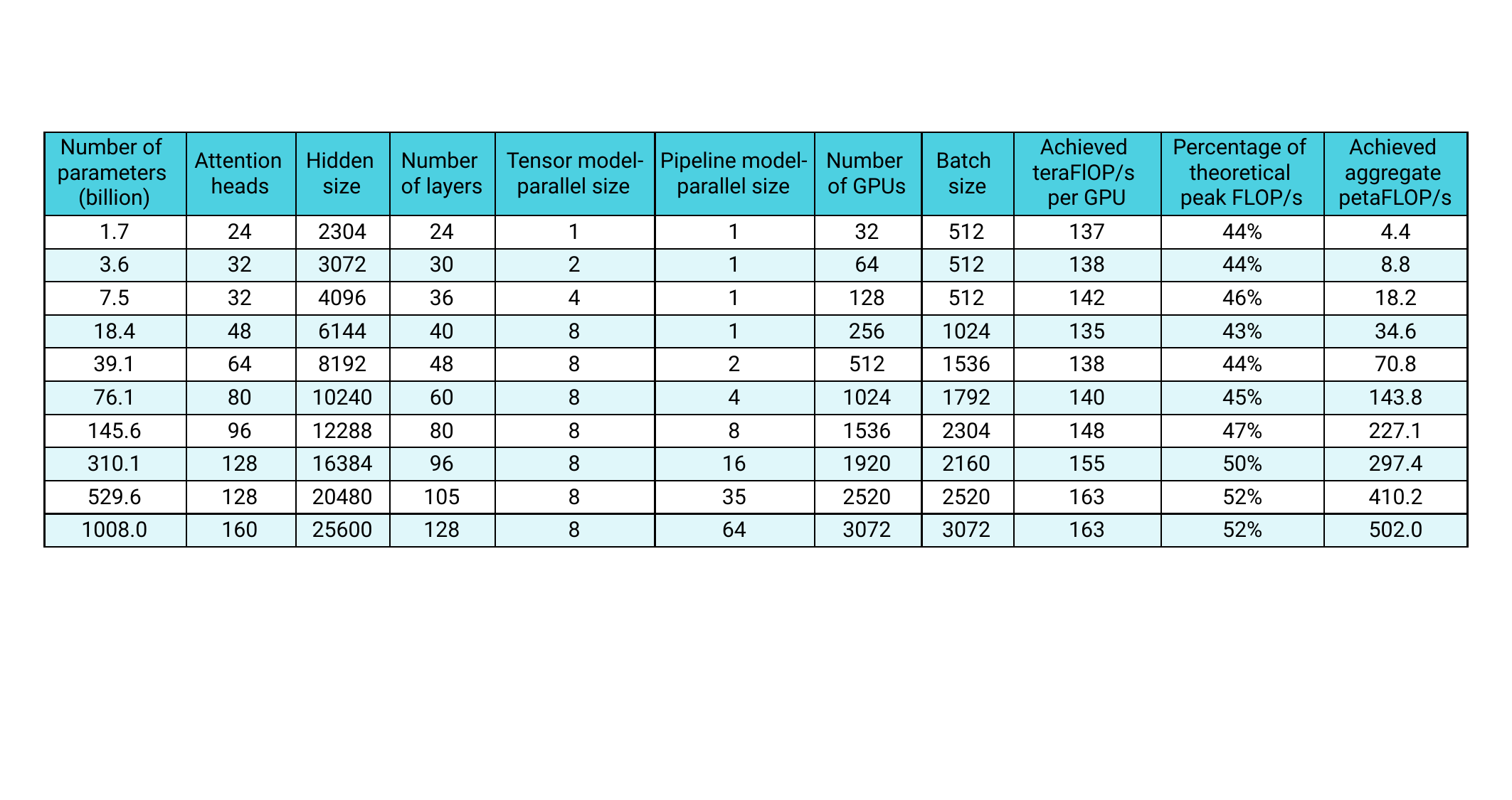}
    \vspace{-1.1in}
    \caption{
         Weak-scaling throughput for GPT models ranging from 1 billion to 1 trillion parameters.
    }
    \label{fig:megatron_scaling}
    \vspace{-0.3in}
\end{table*}

In this section, we seek to answer the following questions:
\begin{itemize}
    \item How well does PTD-P perform? Does it result in realistic end-to-end training times?
    \item How well does pipeline parallelism scale for a given model and batch size? How much impact does the interleaved schedule have on performance?
    \item How do different parallelization dimensions interact with each other? What is the impact of hyperparameters such as microbatch size?
    \item What is the impact of the scatter-gather communication optimization? What types of limits do we put on hardware when running training iterations at scale?
\end{itemize}

All of our results are run with mixed precision on the Selene supercomputer~\cite{selene}. Each cluster node has 8 NVIDIA 80-GB A100 GPUs~\cite{a100}, connected to each other by NVLink and NVSwitch~\cite{nvlink}. Each node has eight NVIDIA Mellanox 200Gbps HDR Infiniband HCAs for application communication, with an additional two HCAs per node for dedicated storage. The nodes are connected in a three-level (leaf, spine, core) fat-tree topology with 850 switches. This topology allows efficient all-reduce communication (dominant communication pattern in deep learning training). The cluster uses an all-NVME shared parallel filesystem for high-performance data access and storage. The peak device throughput of an A100 GPU with 16-bit precision is 312 teraFLOP/s. For most of our results, we report throughput per GPU. Aggregate throughput can be computed by multiplying with the number of GPUs used.

For our experiments, we use GPT models of appropriate sizes. In particular, for any given microbenchmark, the model needs to fit on the number of model-parallel GPUs used in the experiment. We use standard model architectures such as GPT-3~\cite{gpt3} when appropriate.

\subsection{End-to-End Performance}

We consider the end-to-end performance of our system on GPT models ranging from a billion to a trillion parameters, using tensor, pipeline, and data parallelism (degrees picked using heuristics described in \S\ref{sec:parallelization_dimensions}). In particular, we use the interleaved pipeline schedule with the scatter/gather optimization enabled. All models use a vocabulary size (denoted by $V$) of 51,200 (multiple of 1024) and a sequence length ($s$) of 2048. We vary hidden size ($h$), number of attention heads, and number of layers ($l$). The number of parameters in a model, $P$, can be computed as:
\begin{equation}
  P = 12lh^2\left(1 + \dfrac{13}{12h}+\dfrac{V+s}{12lh}\right).
  \label{eq:num-params}
\end{equation}
As the model size increases, we also increase the batch size ($B$) and the number of GPUs ($n$). The majority of floating-point operations in the model are performed in the matrix multiplications (GEMMs) in the transformer and logit layers. Considering just these GEMMs, the number of FLOPs per iteration is (more details in the Appendix):
\begin{equation}
    F=96Bslh^2\left(1 + \dfrac{s}{6h} + \dfrac{V}{16lh}\right).
    \label{eq:flops}
\end{equation}
This is a lower bound for the true FLOP count but should be close to the actual value. We count a FLOP as a floating-point operation regardless of precision. We also note that equation (\ref{eq:flops}) assumes activation recomputation and takes into account the floating-point operations associated with the extra forward pass.

Table~\ref{fig:megatron_scaling} shows the model configurations along with the achieved FLOP/s (both per GPU and aggregate over all GPUs). We see super-linear scaling to 3072 A100 GPUs (384 DGX A100 nodes), since GPU utilization improves as the models get larger (larger matrix multiplications) without significant increase in the communication time relative to computation time. Note that throughput is measured for end-to-end training, i.e., includes all operations including data loading, optimizer steps, communication, and logging. We achieve 52\% of peak device throughput for the largest model, and 44\% of peak device throughput for the smallest model.

\textbf{Training Time Estimates.}
Given these throughputs, we can also estimate the total amount of time needed for end-to-end training on $T$ tokens. Training requires $I=T/\left(B \cdot s\right)$ iterations. Using the value of $F$ from equation (\ref{eq:flops}) and empirical end-to-end throughputs from Table~\ref{fig:megatron_scaling} (denoted by $X$), we can estimate total training time. We note that for the configurations in Table~\ref{fig:megatron_scaling}, we have $6h \gg s$, $16lh \gg \left(V+s\right)$, and $12lh \gg V$. Combining these observations with equations (\ref{eq:num-params}) and  (\ref{eq:flops}), we arrive at
\begin{equation}
\text{End-to-end training time} \approx \dfrac{8TP}{nX}.
\end{equation}
Let us consider the GPT-3 model with $P=$175 billion parameters as an example. This model was trained on $T=300$ billion tokens. On $n=1024$ A100 GPUs using batch size 1536, we achieve $X=140$ teraFLOP/s per GPU. As a result, the time required to train this model is 34 days. For the 1 trillion parameter model, we assume that 450 billion tokens are needed for end-to-end training. With 3072 A100 GPUs, we can achieve a per-GPU throughput of 163 teraFLOP/s, and end-to-end training time of 84 days. We believe these training times (using a reasonable number of GPUs) are practical.

\begin{table*}[t!]
    \vspace{-0.4in}
    \centering
    \includegraphics[keepaspectratio=1.0,width=1.65\columnwidth]{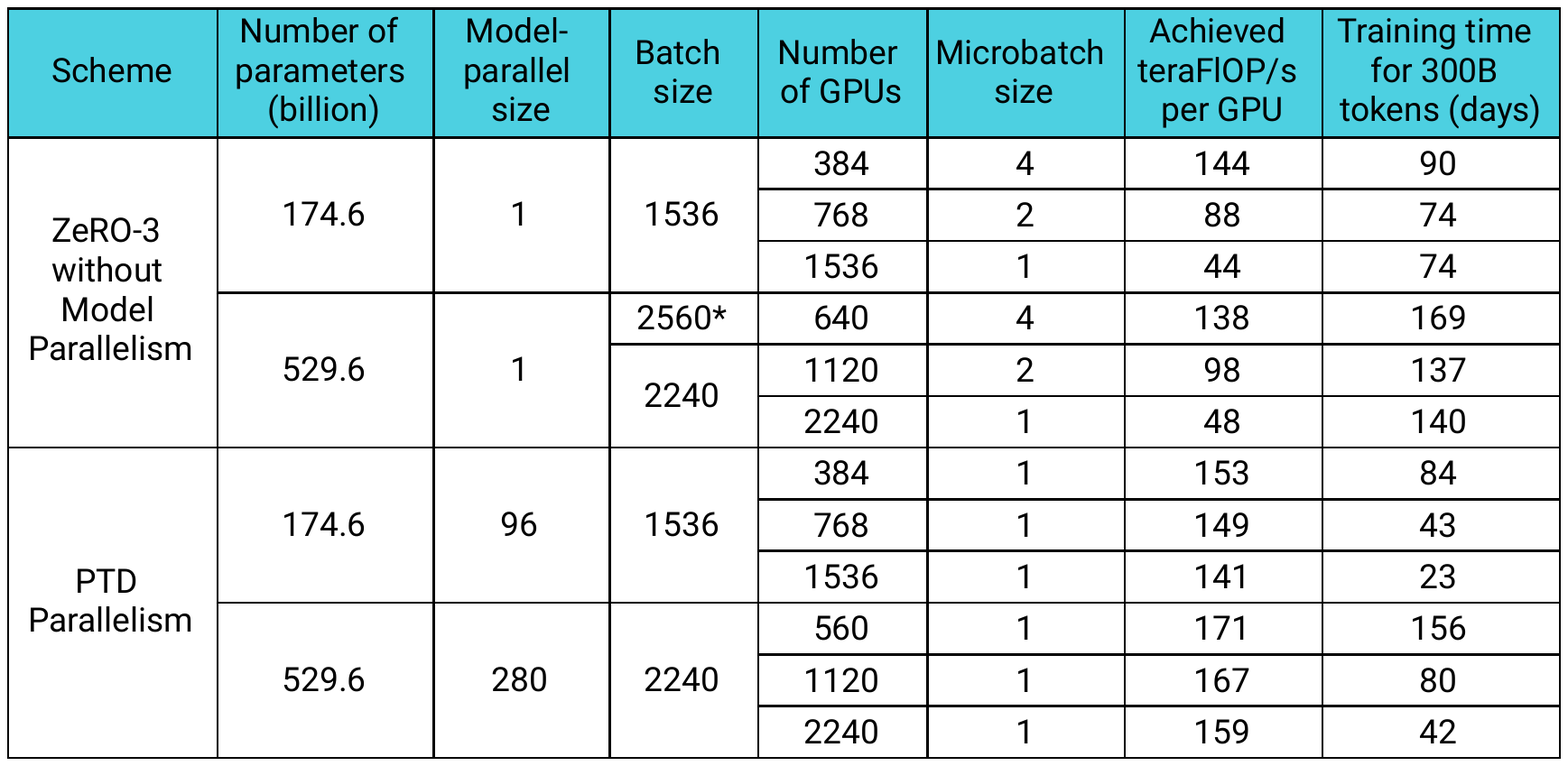}
    \vspace{-4.4in}
    \caption{
         Comparison of PTD Parallelism to ZeRO-3 (without model paralllelism). The 530-billion-parameter GPT model did not fit on 560 GPUs when using a microbatch size of 4 with ZeRO-3, so we increased the number of GPUs used to 640 and global batch size to 2560 to provide a throughput estimate (relevant row marked in table with a *).
    }
    \vspace{-0.25in}
    \label{fig:zero3_comparison_table}
\end{table*}

\begin{figure}[t!]
    \centering
    \includegraphics[keepaspectratio=1.0,width=0.9\columnwidth]{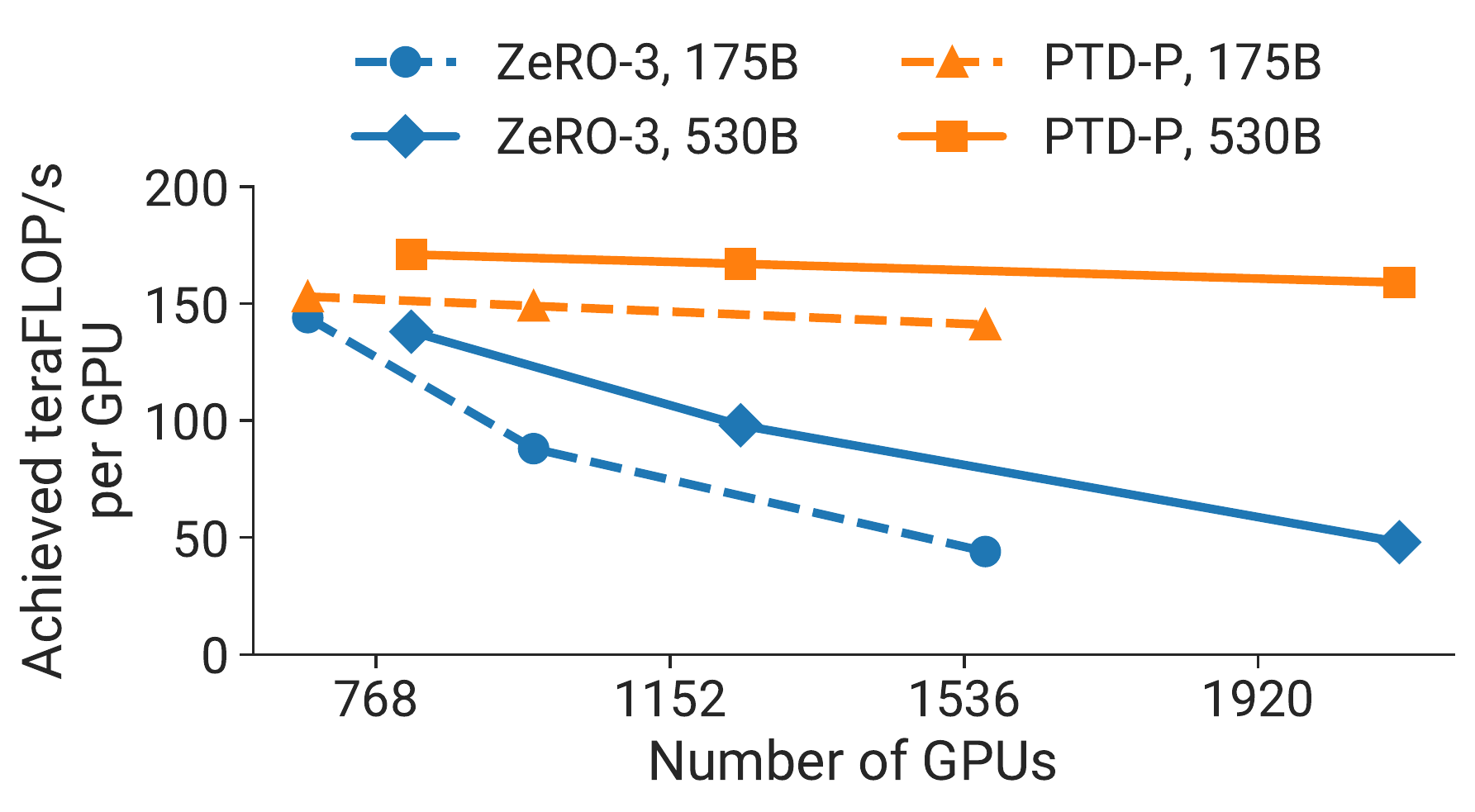}
    \vspace{-0.05in}
    \caption{
         Throughput per GPU of PTD-P and ZeRO-3 for two different GPT models (the 175B GPT-3 model is shown with dotted lines, and the 530B model is shown with solid lines). Global batch sizes are fixed and ZeRO-3 is used without any model parallelism.
    }
    \label{fig:zero3_comparison_throughputs}
    \vspace{-0.2in}
\end{figure}

\subsection{Comparison to ZeRO-3}

We compare PTD-P to ZeRO-3~\cite{rajbhandari2019zero,rajbhandari2021zero} in Table~\ref{fig:zero3_comparison_table} and Figure~\ref{fig:zero3_comparison_throughputs} for the standard GPT-3 model architecture, as well as the 530-billion-parameter model from Table~\ref{fig:megatron_scaling}. The results provide a point of comparison to a method that does not use model parallelism. We integrated ZeRO into our codebase using the \texttt{DeepSpeed} Python library~\cite{deepspeed}. We keep the global batch size the same as we increase the number of GPUs. With fewer GPUs and a microbatch size of 4, PTD-P results in $6\%$ and $24\%$ higher throughput for the 175- and 530-billion-parameter models respectively. As we increase the number of GPUs, PTD-P scales more gracefully than ZeRO-3 in isolation (see Figure~\ref{fig:zero3_comparison_throughputs}). For example, by doubling the number of GPUs (keeping the batch size the same), PTD-P outperforms ZeRO-3 by $70\%$ for both models due to less cross-node communication. We note that we have only considered ZeRO-3 without tensor parallelism. ZeRO-3 can be combined with model parallelism to potentially improve its scaling behavior.

\subsection{Pipeline Parallelism}
We now evaluate the weak-scaling performance of pipeline parallelism in isolation, and also compare the performance of the non-interleaved schedule to the interleaved schedule.

\subsubsection{Weak Scaling}

\begin{figure}[t!]
    \centering
    \includegraphics[keepaspectratio=1.0,width=0.9\columnwidth]{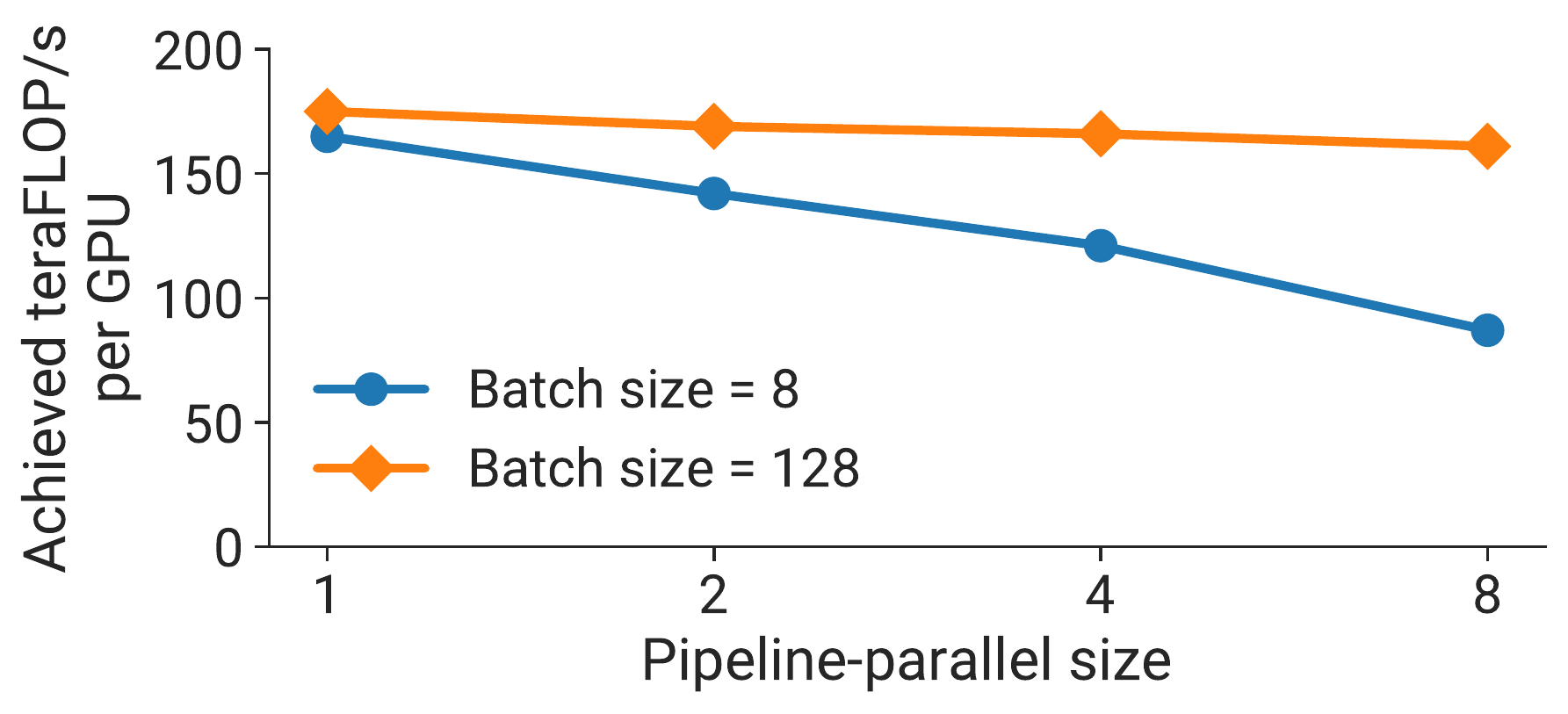}
    \vspace{-0.1in}
    \caption{
         Throughput per GPU of pipeline parallelism using two different batch sizes in a weak-scaling experiment setup (model size increases with the pipeline-parallel size).
    }
    \label{fig:pipeline_parallelism_weak_scaling}
    \vspace{-0.1in}
\end{figure}

We evaluate the scaling of the default non-interleaved pipeline-parallel schedule using a weak scaling setup, a GPT model with 128 attention heads and a hidden size of 20480, and a microbatch size of 1. As we increase the number of pipeline stages, we also increase the size of the model by proportionally increasing the number of layers in the model, e.g., with a pipeline-parallel size of 1, we use a model with 3 transformer layers and ~15 billion parameters, and with a pipeline-parallel size of 8, we use a model with 24 transformer layers and ~121 billion parameters. We use a tensor-parallel size of 8 for all configurations, and vary the total number of A100 GPUs used from 8 to 64. Figure~\ref{fig:pipeline_parallelism_weak_scaling} shows throughput per GPU for two different batch sizes to illustrate the impact of the pipeline bubble, which behaves as $\frac{p-1}{m}$ (\S\ref{sec:pipeline_parallelism_bubble}). As expected, the higher batch size scales better since the pipeline bubble is amortized over more microbatches.

\subsubsection{Interleaved versus Non-Interleaved Schedule}

\begin{figure}[t!]
    \centering
    \includegraphics[keepaspectratio=1.0,width=0.9\columnwidth]{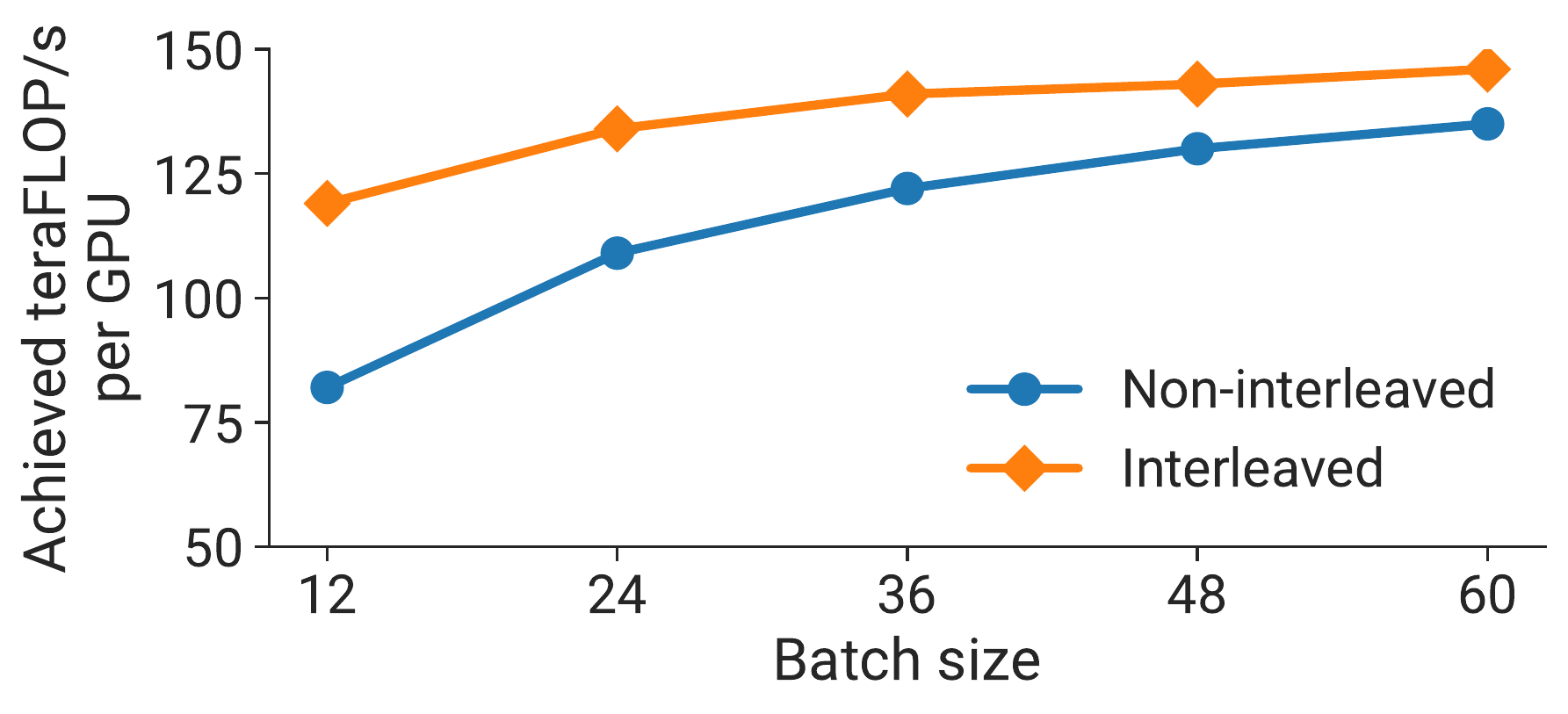}
    \vspace{-0.1in}
    \caption{
         Throughput per GPU of interleaved and non-interleaved schedules for a GPT model (175 billion parameters) on 96 GPUs.
    }
    \vspace{-0.1in}
    \label{fig:interleaved_vs_default}
\end{figure}

Figure~\ref{fig:interleaved_vs_default} shows the per-GPU-throughput for interleaved and non-interleaved schedules on the GPT-3~\cite{gpt3} model with
175 billion parameters (96 layers, 96 attention heads, hidden size of 12288).
The interleaved schedule with the scatter/gather communication optimization has higher computational performance than the non-interleaved (default) schedule. This gap closes as the batch size increases due to two reasons: (a) as the batch size increases, the bubble size in the default schedule decreases, and (b) the amount of point-to-point communication within the pipeline is proportional to the batch size, and consequently the non-interleaved schedule catches up as the amount of communication increases (the interleaved schedule features more communication per sample). Without the scatter/gather optimization, the default schedule performs better than the interleaved schedule at larger batch sizes (not shown).

\subsection{Comparison of Parallel Configurations}
\label{sec:evaluation_parallel_configurations}

In this sub-section, we show the various tradeoffs associated with combining different parallelization dimensions. In particular, we show the performance for parallel configurations using the same number of GPUs for a given model and multiple batch sizes.

\subsubsection{Tensor versus Pipeline Parallelism}

\begin{figure}[t!]
    \centering
    \includegraphics[keepaspectratio=1.0,width=0.9\columnwidth]{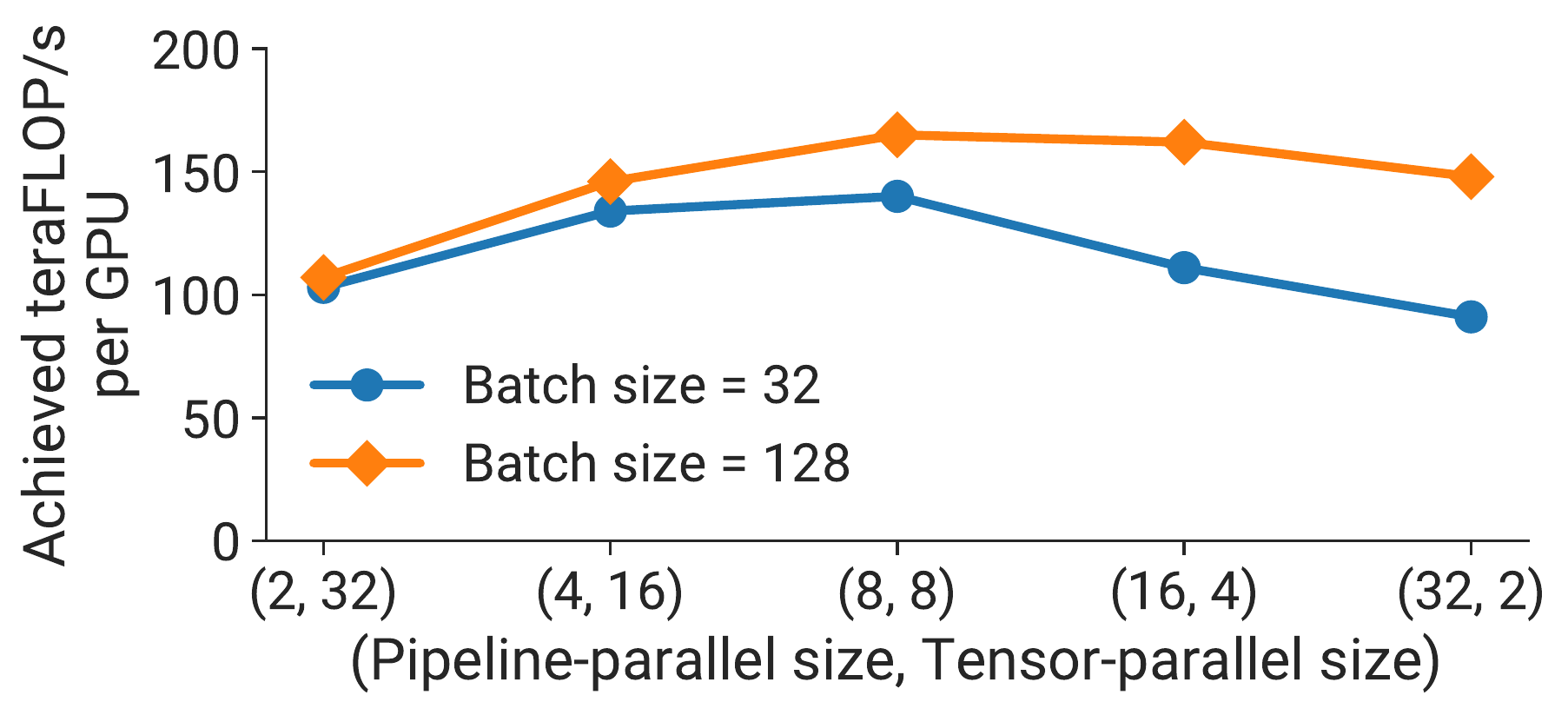}
    \vspace{-0.1in}
    \caption{
         Throughput per GPU of various parallel configurations that combine pipeline and tensor model parallelism using a GPT model with 162.2 billion parameters and 64 A100 GPUs.
    }
    \vspace{-0.1in}
    \label{fig:pipeline_and_tensor_model_parallelism}
\end{figure}

\begin{figure}[t!]
    \centering
    \includegraphics[keepaspectratio=1.0,width=0.9\columnwidth]{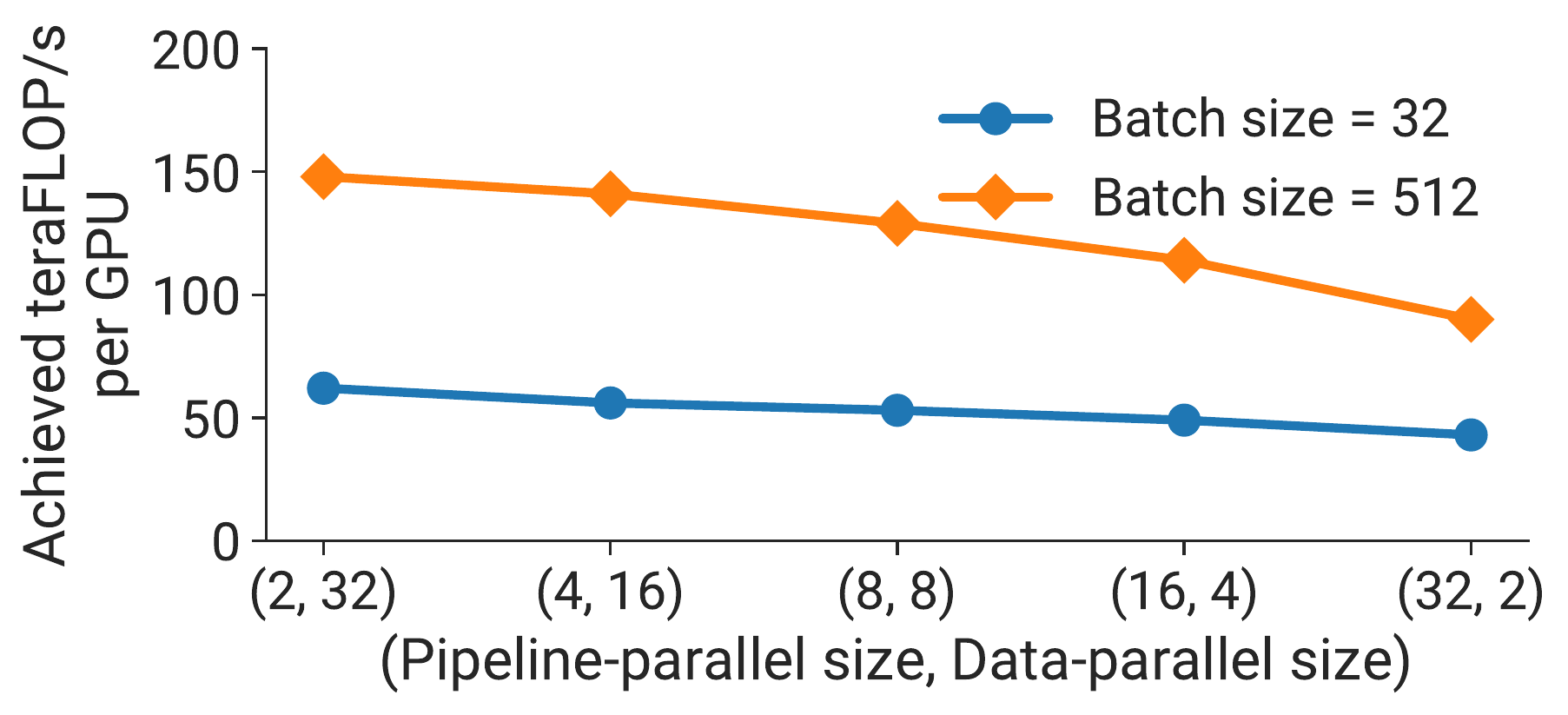}
    \vspace{-0.1in}
    \caption{
         Throughput per GPU of various parallel configurations that combine data and pipeline model parallelism using a GPT model with 5.9 billion parameters, three different batch sizes, microbatch size of 1, and 64 A100 GPUs.
    }
    \vspace{-0.1in}
    \label{fig:data_and_pipeline_model_parallelism}
\end{figure}

\begin{figure}[t!]
    \centering
    \includegraphics[keepaspectratio=1.0,width=0.9\columnwidth]{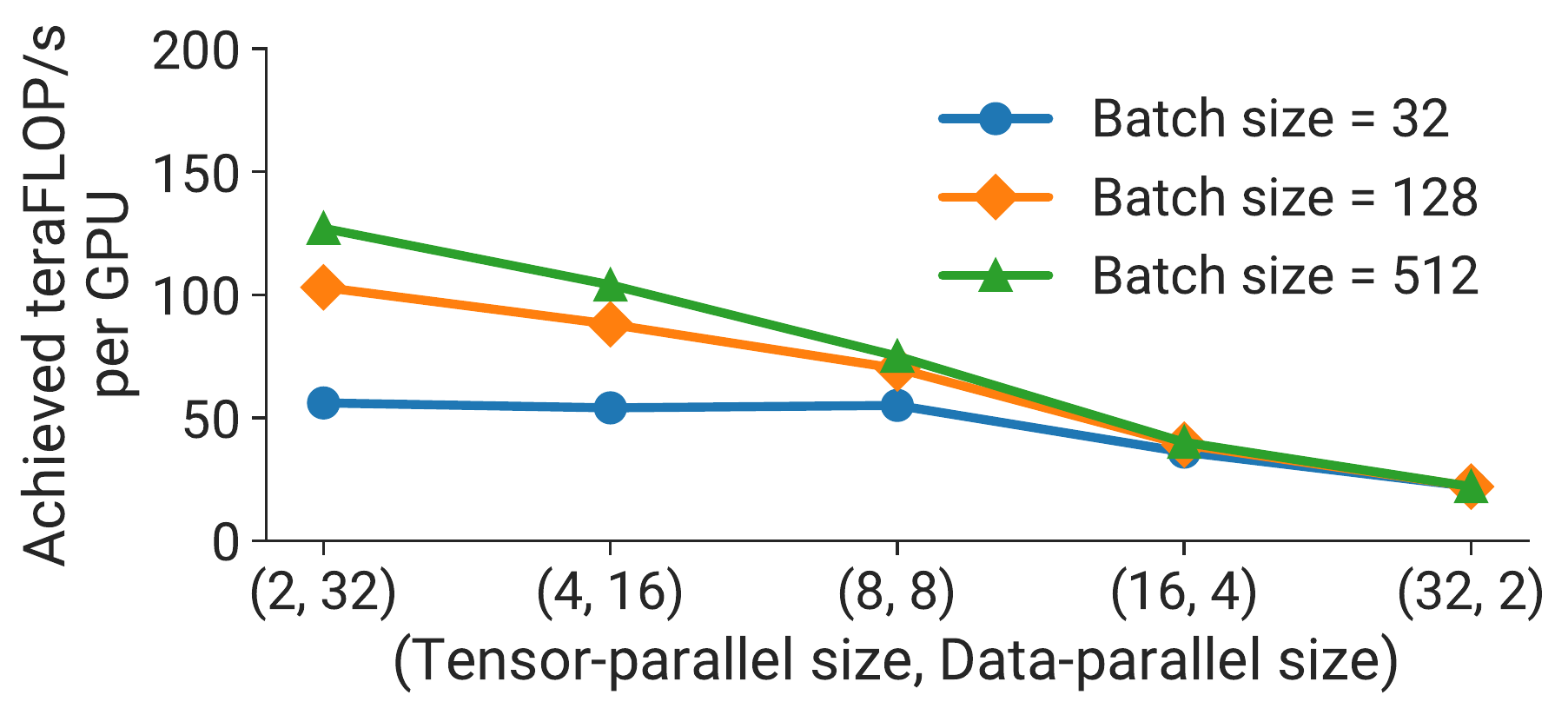}
    \vspace{-0.1in}
    \caption{
         Throughput per GPU of various parallel configurations that combine data and tensor model parallelism using a GPT model with 5.9 billion parameters, three different batch sizes, microbatch size of 1, and 64 A100 GPUs.
    }
    \vspace{-0.1in}
    \label{fig:data_and_tensor_model_parallelism}
\end{figure}

\begin{figure}[t!]
    \centering
    \includegraphics[keepaspectratio=1.0,width=0.9\columnwidth]{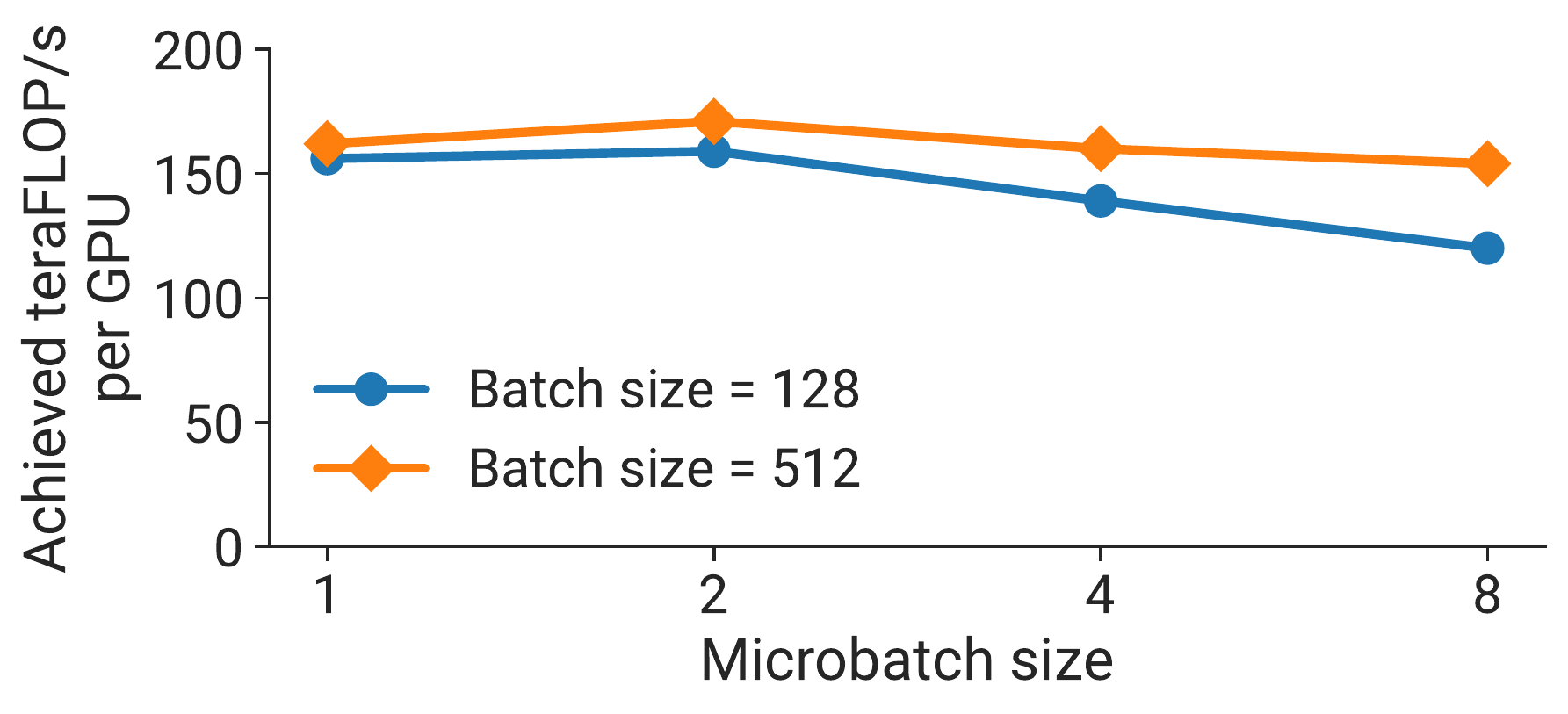}
    \vspace{-0.1in}
    \caption{
         Throughput per GPU of a $(t, p) = (8, 8)$ parallel configuration for different microbatch sizes on a GPT model with 91 billion parameters, for two different batch sizes using 64 A100 GPUs.
    }
    \vspace{-0.1in}
    \label{fig:microbatch_size}
\end{figure}

We evaluate the impact of pipeline and tensor model parallelism on performance for a given model and batch size. The empirical results in Figure~\ref{fig:pipeline_and_tensor_model_parallelism} show the importance of using both tensor and pipeline model parallelism in conjunction to train a 161-billion-parameter GPT model (32 transformer layers to support pipeline-parallel size of 32, 128 attention heads, hidden size of 20480) with low communication overhead and high compute resource utilization. We observe that tensor model parallelism is best within a node (DGX A100 server) due to its expensive all-reduce communication. Pipeline model parallelism, on the other hand, uses much cheaper point-to-point communication that can be performed across nodes without bottlenecking the entire computation. However, with pipeline parallelism, significant time can be spent in the pipeline bubble: the total number of pipeline stages should thus be limited so that the number of microbatches in the pipeline is a reasonable multiple of the number of pipeline stages. Consequently, we see peak performance when the tensor-parallel size is equal to the number of GPUs in a single node (8 with DGX A100 nodes). This result indicates that neither tensor model parallelism (used by Megatron~\cite{shoeybi2019megatron}) nor pipeline model parallelism (used by PipeDream~\cite{narayanan2021memory} and others) in isolation can match the performance of using both techniques in conjunction.

\subsubsection{Pipeline versus Data Parallelism}

We evaluate the impact of data and pipeline model parallelism on performance for a GPT model with 5.9 billion parameters (32 transformer layers, 32 attention heads, hidden size of 3840) in Figure~\ref{fig:data_and_pipeline_model_parallelism}. We use a smaller model than before since we want to show performance for models that fit when the model-parallel size is only 2. For simplicity, we keep the microbatch size equal to 1 in these experiments. We see that for each batch size, the throughput decreases as the pipeline-parallel size increases, matching our analytical model from \S\ref{sec:data_and_model_parallelism_analytical_model}. Pipeline model parallelism should be used primarily to support the training of large models that do not fit on a single worker, and data parallelism should be used to scale up training.

\subsubsection{Tensor versus Data Parallelism}

We also evaluate the impact of data and tensor model parallelism on performance for the same GPT model with 5.9 billion parameters in Figure~\ref{fig:data_and_tensor_model_parallelism} (smaller model used for same reason as above). As before, we keep the microbatch size equal to 1 initially. With larger batch sizes and a microbatch size of 1, data-parallel communication is infrequent; the all-to-all communication required in tensor model parallelism needs to be performed for \emph{every} microbatch in a batch. This all-to-all communication with tensor model parallelism dominates end-to-end training time, especially when communication needs to be performed across multi-GPU nodes. Additionally, as the tensor-model-parallel size increases, we perform smaller matrix multiplications on every GPU, decreasing utilization on each GPU.

\vspace{0.1in}
We should note that although data parallelism can lead to efficient scaling, we cannot use data parallelism in isolation for very large models with a limited training batch size because of a) insufficient memory capacity, and b) scaling limitations of data parallelism (e.g., GPT-3 was trained to convergence with a batch size of $1536$. Data parallelism thus supports parallelization to only $1536$ GPUs; however, roughly $10,000$ GPUs were used to train this model in a reasonable amount of time).

\subsection{Microbatch Size}

We evaluate the impact of the microbatch size on the performance of parallel configurations that combine pipeline and tensor model parallelism in Figure~\ref{fig:microbatch_size} for a model with 91 billion parameters ($(t, p) = (8, 8)$). We see that the best microbatch size is 2 for this model; the optimal microbatch size is different for other models (not shown in Figure) and \emph{model-dependent}. For a given batch size, increasing the microbatch size decreases the number of microbatches in the pipeline ($m$), leading to a larger pipeline bubble; however, increasing the microbatch size can also improve GPU utilization by increasing the arithmetic intensity of executed kernels. These two factors are at odds with each other, which makes the choice of optimal microbatch size challenging. Our analytical model from \S\ref{sec:data_and_model_parallelism_analytical_model} reasonably approximates true performance, and can be used as a proxy to determine how to pick this hyperparameter value for various training configurations and models.

\begin{figure}[t!]
    \centering
    \includegraphics[keepaspectratio=1.0,width=0.9\columnwidth]{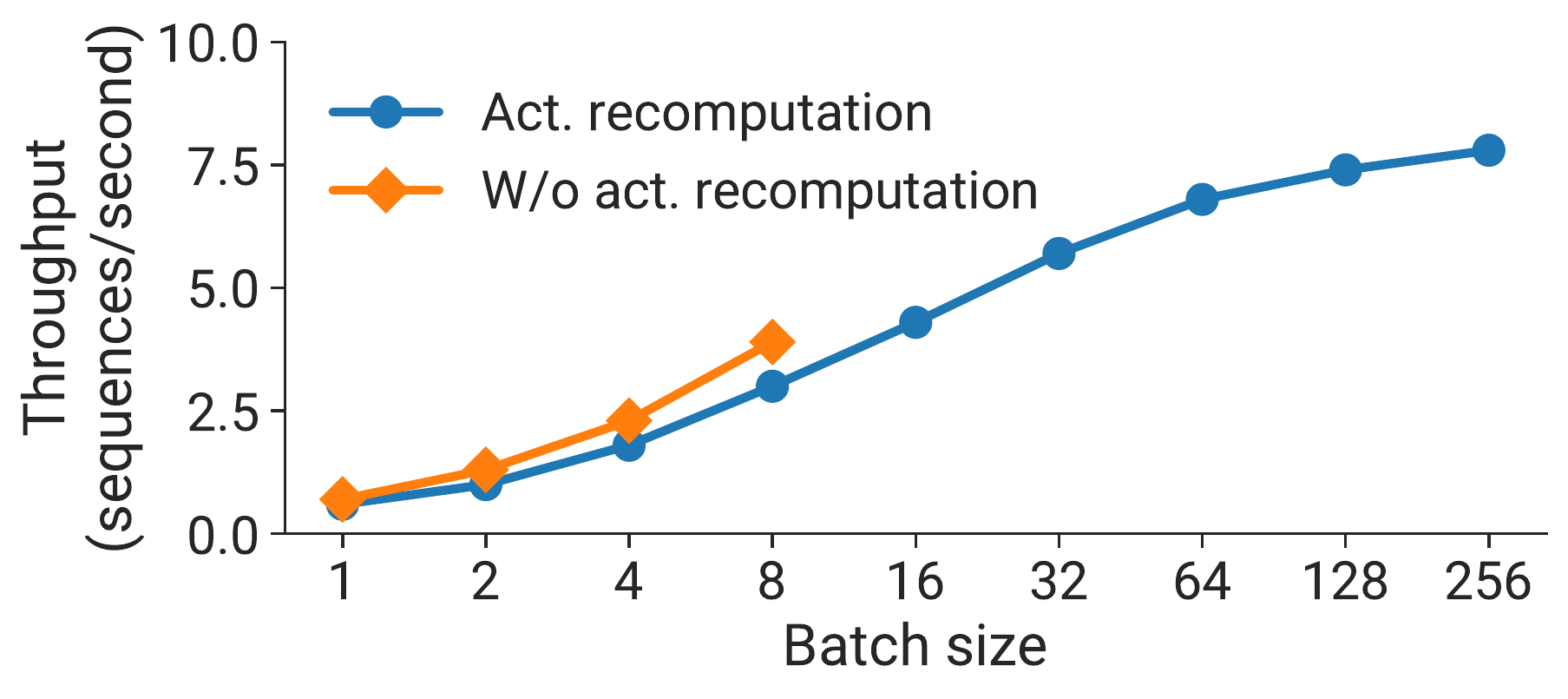}
    \vspace{-0.1in}
    \caption{
         Throughput (in sequences per second) with and without activation recomputation for a GPT model with 145 billion parameters using 128 A100 GPUs ($(t, p) = (8, 16)$).
    }
    \vspace{-0.1in}
    \label{fig:activation_recomputation}
\end{figure}

\subsection{Activation Recomputation}

Figure~\ref{fig:activation_recomputation} shows throughput with and without activation recomputation for a GPT model with 145 billion parameters (80 transformer layers, 96 attention heads, hidden size of 12288) using 128 A100 GPUs, $(t, p) = (8, 16)$, and a range of batch sizes. For small batch sizes, activation recomputation leads to up to $33\%$ lower throughput (in sequences per second) due to the extra forward pass that needs to be executed during the backward pass. However, activation recomputation is needed to support larger batch sizes. Throughput at large batch sizes with activation recomputation is up to $2\times$ higher than the best throughput achieved without activation recomputation (for a smaller batch size) due to a smaller pipeline bubble.

\begin{figure}[t!]
    \centering
    \includegraphics[keepaspectratio=1.0,width=0.9\columnwidth]{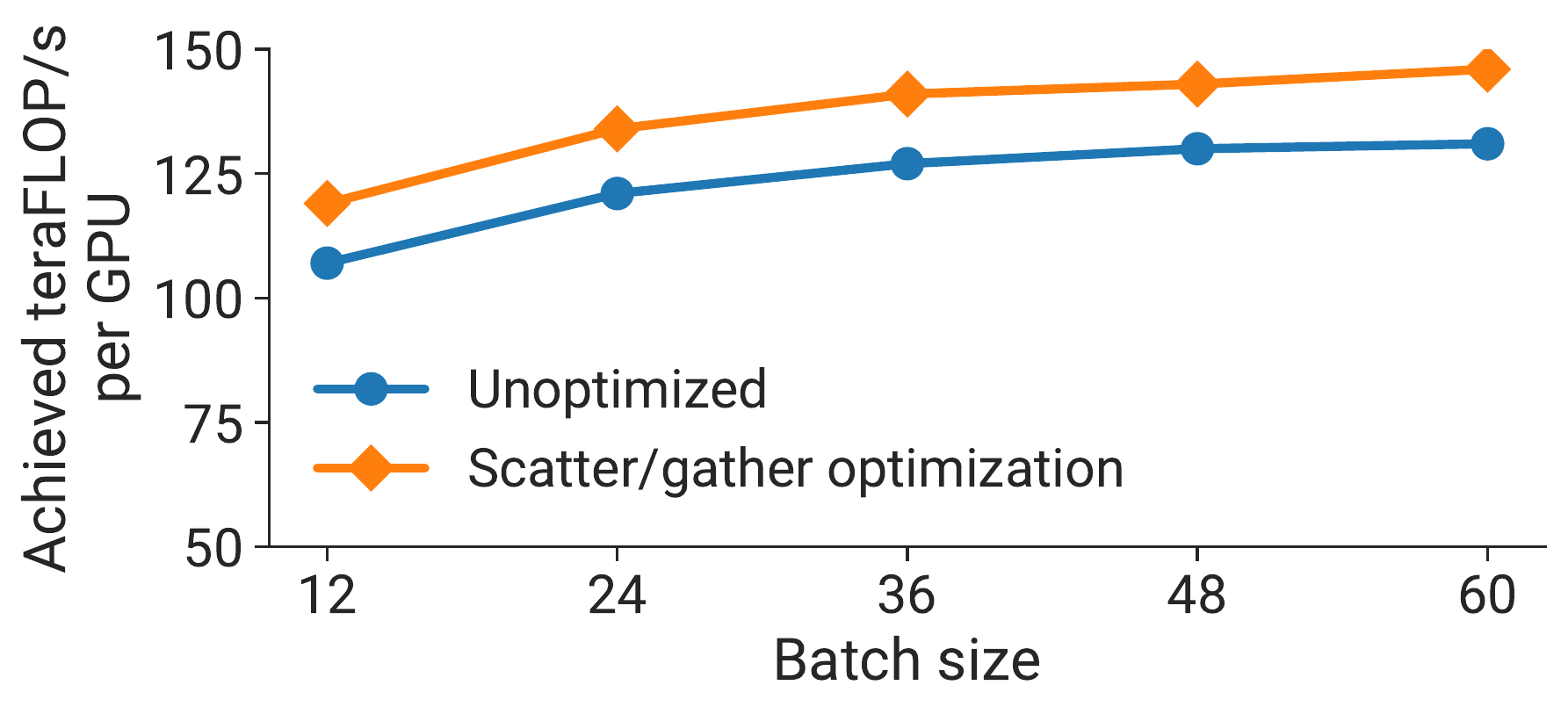}
    \vspace{-0.1in}
    \caption{
         Throughput per GPU with and without the scatter/gather optimization for a GPT model with 175 billion parameters using 96 A100 GPUs and the interleaved schedule.
    }
    \vspace{-0.1in}
    \label{fig:scatter_gather_optimization_eval}
\end{figure}

\subsection{Scatter-Gather Optimization}

Figure~\ref{fig:scatter_gather_optimization_eval} shows per-GPU-throughput with and without (unoptimized) the scatter/gather communication optimization for the GPT-3 model with 175 billion parameters. We see an improvement of up to 11\% in throughput for communication-intensive schedules (large batch size with interleaving) by reducing the amount of communication over cross-node links.

\subsection{Fused Operators}
We also evaluate the performance impact of operator fusion described in \S\ref{sec:computation_optimizations}. For the GPT-3 model (175 billion parameters), throughput increased by 19\% with fusion (113 teraFLOP/s per GPU to 135 teraFLOP/s per GPU). For the larger GPT model with 530 billion parameters (model configuration in Figure~\ref{fig:megatron_scaling}), throughput increased by 11\% (133 teraFLOP/s per GPU to 148 teraFLOP/s per GPU).

\subsection{Inter-Node Communication Bandwidth}
Our strong results are a byproduct of using an optimized software and hardware stack \emph{together}. In particular, we take advantage of the high-bandwidth communication links between GPUs on the same server and across servers. On the trillion-parameter model with 3072 GPUs, we observed that the effective bisection bandwidth of point-to-point communication among pipeline stages is 892 GB/s, while the effective bisection bandwidth of all-reduce operations among data-parallel replicas is 12.9 TB/s. A less-optimized partitioning of operators across devices would lead to more inter-node communication, hampering scaling performance.

\subsection{Checkpoint Loading and Saving}

An important practical consideration for the training of large models is loading and saving model checkpoints, which are especially large for the models considered in this paper. For example, the trillion-parameter model has a checkpoint of size 13.8 terabytes. The initial load of checkpoints for the trillion-parameter model by all 384 nodes (3072 GPUs) reaches a peak read bandwidth of 1TB/s, the maximum read throughput possible from the parallel filesystem. Checkpoint saves reach 40\% of peak write bandwidth (273 GB/s).
\section{Related Work}

In this section, we discuss other techniques to train models at scale.

\paragraph{Parallelism for Large Models.} Pipeline model parallelism is a common technique used to train large models. Pipeline parallelism comes in a few flavors: the mode discussed in this paper uses flushes to ensure \emph{strict} optimizer semantics. TeraPipe~\cite{li2021terapipe} exposes fine-grained pipeline parallelism across tokens in a single training sequence for auto-regressive models like GPT. PipeTransformer~\cite{he2021pipetransformer} elastically adjusts the degree of pipelining and data parallelism by freezing layers with ``stable'' weights, and instead dedicates resources to train the remaining ``active'' layers. HetPipe~\cite{park2020hetpipe} uses a combination of pipeline and data parallelism on a set of heterogeneous accelerators. Pipeline parallelism can also be implemented with relaxed semantics: PipeDream-\texttt{2BW}~\cite{narayanan2021memory} maintains two weight versions and guarantees 1-stale weight updates without expensive flushes, while PipeMare~\cite{yang2021pipemare} and Kosson et al.~\cite{pb2021kosson} use asynchoronous pipeline parallelism. These techniques have improved throughput compared to the techniques with pipeline flushes considered in this paper, but potentially at the cost of convergence rate or final accuracy. Moreover, pipeline parallelism in isolation can still only scale to a number of devices equal to the number of layers in the model, which is limiting for certain model architectures.

PipeDream~\cite{narayanan2019pipedream} combined pipeline parallelism and data parallelism in a principled way to reduce cross-device communication. DeepSpeed~\cite{deepspeed3Dparallelism} combined pipeline parallelism with tensor and data parallelism to train models with up to a trillion parameters, but with lower throughput than what was shown in this paper (52\% vs. 36\% of peak) for a few reasons: operator fusion to keep most of the operator graph compute-bound, a more-efficient pipeline parallelism schedule to minimize the pipeline bubble size, fast hardware (A100 vs. V100 GPUs and high-bandwidth links between GPUs on the same and different servers), and scaling to more GPUs. We want to emphasize that this higher throughput makes estimated training times much more practical (about 3 months); an aggregate throughput of 37.6 petaFLOP/s would take about 40 months to train an equivalently-sized model. We can scale to larger models as well, but would need more GPUs to keep training time practical.

Mesh-TensorFlow~\cite{mesh_tf} proposes a language for easily specifying parallelization strategies that combine data and model parallelism. Switch Transformers~\cite{fedus2021switch} used Mesh-Tensorflow to train a sparsely activated expert-based model with 1.6 trillion parameters, with improved pre-training speed over the T5-11B model~\cite{T5}.

\paragraph{Sharded Data Parallelism.} As part of performance optimizations for MLPerf 0.6~\cite{mattson2019mlperf}, sharded data parallelism~\cite{kumar2019scale, xu2020automatic}, where optimizer state is sharded over data-parallel workers, was introduced. This method has two advantages: (a) it does not introduce extra communication over vanilla data parallelism, and (b) it divides the optimizer's computation and memory cost across the data-parallel partitions. ZeRO~\cite{rajbhandari2019zero,rajbhandari2021zero} extends this idea: weight parameters and gradients are sharded across data-parallel workers as well, and workers fetch relevant state from their ``owning'' workers before performing computations. This adds additional communication, which can be partially hidden by carefully overlapping computation and communication. However, this can become harder if tensor parallelism is not used or the batch size is not large enough to hide the extra communication overhead (Figure~\ref{fig:zero3_comparison_throughputs}). ZeRO-Infinity~\cite{rajbhandari2021zero} uses NVMe to efficiently swap parameters, enabling the training of very large models on a small number of GPUs. We note that using a small number of GPUs for training a very large model results in unrealistic training times (e.g., thousands of years to converge).

\paragraph{Automatic Partitioning.} FlexFlow~\cite{flexflow}, PipeDream~\cite{narayanan2019pipedream}, DAPPLE~\cite{fan2021dapple}, and Tarnawski et al.~\cite{tarnawski2020efficient} all auto-partition model training graphs over multiple devices with the help of cost models. However, each of these do not consider \emph{all} the parallelism dimensions considered in this paper: pipeline and tensor model parallelism, data parallelism, microbatch size, and the effect of memory-savings optimizations like activation recomputation on the training of models larger than the memory capacity of an accelerator. These added dimensions increase the search space that needs to be explored. Gholami et al.~\cite{gholami2018integrated} show how communication costs for combinations of data and model parallelism can be modeled.

\paragraph{HPC for Model Training.} Goyal et al.~\cite{goyal2017accurate} and You et al.~\cite{you2018imagenet} both demonstrate the use of High Performance Computing techniques to train highly-accurate ImageNet models in minutes. However, the image classification models considered fit comfortably on a single accelerator, rendering model parallelism unnecessary, support very large batch sizes ($>32$k) that allow scaling data parallelism to large worker counts with infrequent communication, and are composed of compact convolutional layers that are inherently amenable to data-parallel communication.

\section{Discussion and Conclusion}

In this paper, we have shown how PTD-P (inter-node pipeline parallelism, intra-node tensor parallelism, and data parallelism) can be composed to achieve high aggregate throughput (502 petaFLOP/s) while training large models with a trillion parameters. This facilitates end-to-end training in reasonable times (estimated time of around 3 months for a trillion-parameter model). We discussed the various tradeoffs associated with each of these types of parallelism, and how the interactions between them need to be considered carefully when combined.

Even though the implementation and evaluation in this paper is GPU-centric, many of these ideas translate to other types of accelerators as well. Concretely, the following are ideas that are accelerator-agnostic: a) the idea of smartly partitioning the model training graph to minimize the amount of communication while still keeping devices active, b) minimizing the number of memory-bound kernels with operator fusion and careful data layout, c) other domain-specific optimizations (e.g., scatter-gather optimization).
\section*{Acknowledgements}

We thank the anonymous reviewers, Seonmyeong Bak, Keshav Santhanam, Trevor
Gale, Dimitrios Vytiniotis, and Siddharth Karamcheti for their help and feedback
that improved this work. This research was supported in part by NSF Graduate
Research Fellowship grant DGE-1656518 and NSF CAREER grant CNS-1651570. Any
opinions, findings, and conclusions or recommendations expressed in this material
are those of the authors alone.
\appendix

\section*{Appendix: Floating-Point Operations}
\label{sec:appendix}

In this section, we describe how we calculate the number of floating-point operations (FLOPs) in a model. We consider a language model with $l$ transformer layers, hidden size $h$, sequence length $s$, vocabulary size $V$, and training batch size $B$.

A $A_{m \times k} \times X_{k \times n}$ matrix multiplication requires $2m \times k \times n$ FLOPs (factor of 2 needed to account for multiplies and adds). 

A transformer layer consists of an attention block followed by a 2-layer feed-forward network. For the attention block, the main FLOP contributors are the key, query, and value transformation ($6Bsh^2$ operations), attention matrix computation ($2Bs^2h$ operations), attention over values ($2Bs^2h$ operations), and post-attention linear projection ($2Bsh^2$ operations). The feed-forward network increases the hidden size to $4h$ and then reduces it back to $h$; this requires $16Bsh^2$ FLOPs. Summing these together, each transformer layer results in $24Bsh^2 + 4Bs^2h$ FLOPs for the forward pass. The backward pass requires double the number of FLOPs since we need to calculate the gradients with respect to both input and weight tensors. In addition, we are using activation recomputation, which requires an additional forward pass before the backward pass. As a result, the total number of FLOPs per transformer layer is
$4 \times \left(24Bsh^2 + 4Bs^2h\right) = 96Bsh^2\left(1 + \dfrac{s}{6h}\right)$.

The other main contributor to the FLOP count is the logit layer in the language model head, which transforms features of dimension $h$ to the vocabulary dimension $V$. The required FLOPs for this operation is $2BshV$ in the forward pass and $4BshV$ in the backward pass, resulting in $6BshV$ FLOPs in total.

Thus, for a transformer model with $l$ transformer layers, the total number of floating-point operations is:
$$96Bslh^2\left(1 + \dfrac{s}{6h} + \dfrac{V}{16lh}\right).$$

\newpage

\bibliography{paper}
\bibliographystyle{plain}

\end{document}